\documentclass{article}

\usepackage{microtype}
\usepackage{graphicx}
\usepackage{subfigure}
\usepackage{booktabs} %

\usepackage{hyperref}

\usepackage[accepted]{icml2025}

\usepackage{amsmath}
\usepackage{amssymb}
\usepackage{mathtools}
\usepackage{amsthm}

\usepackage[capitalize,noabbrev]{cleveref}
\usepackage{color,xcolor}
\usepackage{epsfig}
\usepackage{graphicx}

\usepackage{cuted}
\usepackage{adjustbox}
\usepackage{array}
\usepackage{booktabs}
\usepackage{colortbl}
\usepackage{float,wrapfig}
\usepackage{hhline}
\usepackage{multirow}
\usepackage{subcaption} %

\usepackage{amsmath,amsfonts}
\usepackage{bm}
\usepackage{nicefrac}
\usepackage{microtype}
\usepackage{inconsolata}
\usepackage{pifont}

\usepackage{changepage}
\usepackage{extramarks}
\usepackage{fancyhdr}
\usepackage{lastpage}
\usepackage{setspace}
\usepackage{soul}
\usepackage{xspace}
\usepackage{indentfirst}
\usepackage{paralist}
\usepackage{dcolumn,booktabs}
\newcolumntype{d}[1]{D{.}{.}{#1}}

\usepackage{url}

\usepackage{letltxmacro}

\definecolor{citecolor}{HTML}{0071bc}
\definecolor{mydarkblue}{rgb}{0,0.08,1}
\definecolor{mydarkgreen}{rgb}{0.02,0.6,0.02}
\definecolor{mydarkred}{rgb}{0.8,0.02,0.02}
\definecolor{mydarkorange}{rgb}{0.40,0.2,0.02}
\definecolor{mypurple}{RGB}{111,0,255}
\definecolor{myred}{rgb}{1.0,0.0,0.0}
\definecolor{mygold}{rgb}{0.75,0.6,0.12}
\definecolor{mydarkgray}{rgb}{0.66, 0.66, 0.66}

\definecolor{darkblue}{rgb}{0,0.08,1}
\definecolor{darkgreen}{rgb}{0.02,0.6,0.02}
\definecolor{darkred}{rgb}{0.8,0.02,0.02}
\definecolor{darkorange}{rgb}{0.40,0.2,0.02}
\definecolor{darkpurple}{RGB}{111,0,255}

\newcommand{\todocite}[1]{{\color{blue}{[citation needed]}}}

\newcolumntype{H}{>{\setbox0=\hbox\bgroup}c<{\egroup}@{}}

\definecolor{purple}{RGB}{160, 32, 240}
\definecolor{washblue}{RGB}{186, 224, 228}
\definecolor{sky}{RGB}{128, 128, 128}
\definecolor{seagreen}{RGB}{60, 179, 113}
\definecolor{building}{RGB}{128, 0, 0}
\definecolor{road}{RGB}{128, 64, 128}
\definecolor{sidewalk}{RGB}{0, 0, 192}
\definecolor{fence}{RGB}{64, 64, 128}
\definecolor{vegetation}{RGB}{128, 128, 0}
\definecolor{car}{RGB}{64, 0, 128}
\definecolor{sign}{RGB}{192, 128, 128}
\definecolor{pedestrian}{RGB}{64, 64, 0}
\definecolor{cyclist}{RGB}{0, 128, 192}

\def\be {\begin{equation}}
\def\ee {\end{equation}}
\def\beas {\begin{eqnarray*}}
\def\eeas {\end{eqnarray*}}
\def\bea {\begin{eqnarray}}
\def\eea {\end{eqnarray}}
\def\bes {\begin{equation*}}
\def\ees {\end{equation*}}

\usepackage{amssymb}%
\usepackage{pifont}%

\makeatletter
\usepackage{xspace}
\def\@onedot{\ifx\@let@token.\else.\null\fi\xspace}
\DeclareRobustCommand\onedot{\futurelet\@let@token\@onedot}

\def\ie{\emph{i.e}\onedot}

\makeatother

\usepackage[utf8]{inputenc} %
\usepackage[T1]{fontenc}    %
\usepackage{hyperref}       %
\usepackage{url}            %
\usepackage{booktabs}       %
\usepackage{amsfonts}       %
\usepackage{nicefrac}       %
\usepackage{microtype}      %
\usepackage{xcolor}         %
\usepackage{algorithm, algpseudocode}
\usepackage{amsmath}
\usepackage{amsfonts}
\usepackage{float}

\theoremstyle{plain}

\theoremstyle{definition}

\theoremstyle{remark}

\usepackage[textsize=tiny]{todonotes}

\icmltitlerunning{Introducing 3D Representation for Medical Image Volume-to-Volume Translation via Score Fusion}

\begin{document}

\twocolumn[
\icmltitle{Introducing 3D Representation for Medical Image Volume-to-Volume Translation via Score Fusion}

\icmlsetsymbol{equal}{*}

\begin{icmlauthorlist}
\icmlauthor{Xiyue Zhu}{UIUC}
\icmlauthor{Dou Hoon Kwark}{UIUC}
\icmlauthor{Ruike Zhu}{UIUC}
\icmlauthor{Kaiwen Hong}{UIUC}
\icmlauthor{Yiqi Tao}{UIUC}
\icmlauthor{Shirui Luo}{NCSA}
\icmlauthor{Yudu Li}{UIUC}
\icmlauthor{Zhi-Pei Liang}{UIUC}
\icmlauthor{Volodymyr Kindratenko$^{1,2}$}{}
\end{icmlauthorlist}
\hspace{13mm} \textsuperscript{1}University of Illinois at Urbana-Champaign \hspace{5mm} \textsuperscript{2} National Center for Supercomputing Applications

\icmlaffiliation{UIUC}{University of Illinois at Urbana-Champaign}
\icmlaffiliation{NCSA}{National Center for Supercomputing Applications}

\icmlcorrespondingauthor{Xiyue Zhu}{xiyuez2@illinois.edu}

\icmlkeywords{Generative Models, Diffusion Models, 3D Medical image}

\vskip 0.3in
]

\begin{abstract}
\label{sec:Abstract}

In volume-to-volume translations in medical images, existing models often struggle to capture the inherent volumetric distribution using 3D voxel-space representations, due to high computational dataset demands. We present Score-Fusion, a novel volumetric translation model that effectively learns 3D representations by ensembling perpendicularly trained 2D diffusion models in score function space. By carefully initializing our model to start with an average of 2D models as in TPDM \cite{lee2023improving}, we reduce 3D training to a fine-tuning process and thereby mitigate both computational and data demands. Furthermore, we explicitly design the 3D model's hierarchical layers to learn ensembles of 2D features, further enhancing efficiency and performance. Moreover, Score-Fusion naturally extends to multi-modality settings, by fusing diffusion models conditioned on different inputs for flexible, accurate integration. We demonstrate that 3D representation is essential for better performance in downstream recognition tasks, such as tumor segmentation, where most segmentation models are based on 3D representation. Extensive experiments demonstrate that Score-Fusion achieves superior accuracy and volumetric fidelity in 3D medical image super-resolution and modality translation. Beyond these improvements, our work also provides broader insight into learning-based approaches for score function fusion.

\end{abstract}

\section{Introduction}

\begin{figure}[t] \centering 
\includegraphics[width=0.5\textwidth]{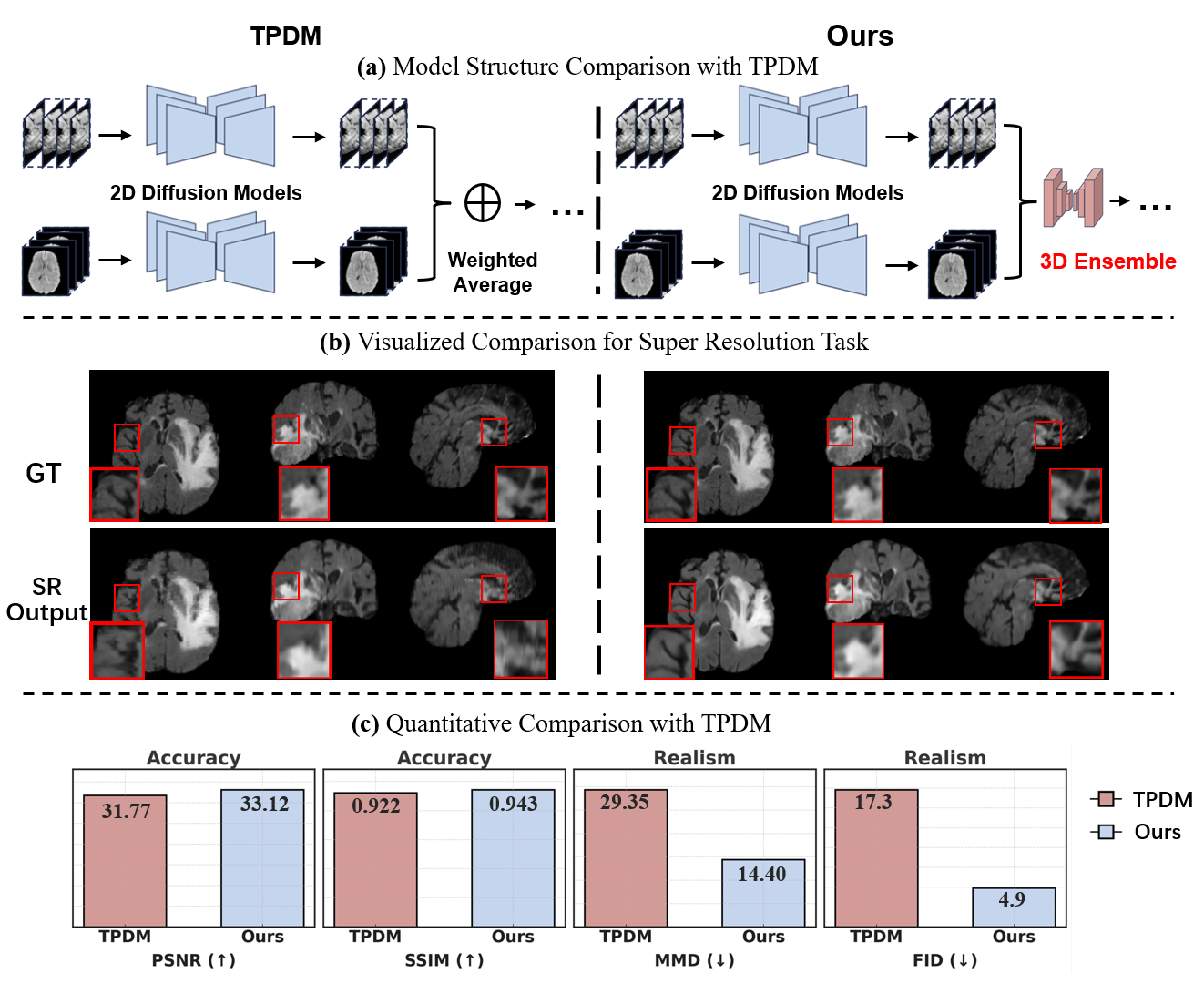} 
\caption{Comparison between TPDM (left) and Score-Fusion (right). Score-Fusion learns to ensemble pre-trained diffusion models with a 3D model, effectively utilizing 3D representations. Our model thus shows better 3D realism and demonstrates superior accuracy and realism metrics. }
\label{fig:intro}
\end{figure}

\label{sec:Introduction}

Volume-to-volume image translation is critical for volumetric medical imaging, such as magnetic resonance imaging (MRI) and X-ray computed tomography (CT). It addresses various inverse problems of image reconstruction~\cite{Hai-Miao, article, article2}, handling sparse~\cite{doi:10.1002/mrm.26977, mardani2017deep}, limited~\cite{9941138, 8950304}, and/or noisy~\cite{9941138, article3} imaging data. It also supports image synthesis, such as multi-contrast MRI~\cite{article4, Dar2018ImageSI,shi2021frequency, wolterink2017deep}, CT-ultrasound~\cite{vedula2017towards}, and MR-histopathology~\cite{pmlr-v156-leroy21a}.

3D voxel-space representation plays a critical role in volume-to-volume translation tasks for various reasons. Firstly, medical images are inherently 3D-dense volumes. Using a 3D representation enables us to directly model the entire 3D distribution. Additionally, the 3D latent diffusion model (LDM) is a common technique to accelerate diffusion models. However, compared to LDMs, the 3D voxel-space diffusion model maintains better features in high-frequency details, which are essential in inverse problems that require highly accurate prediction, such as super-resolution. Moreover, volumetric translation models enable numerous downstream image processing tasks, ranging from image reconstruction~\cite{1ea325989ead4045be152d74394d539a} to analysis~\cite{akrout2023diffusion, fernandez2022can}). Most models in analysis tasks, such as tumor segmentation\cite{SWINUnet}, are trained with 3D volumes using voxel-space 3D representations. Therefore, volume translation models using 3D representation may generate images that are more accurate when processed with such downstream task models.

However, previous works~\cite{dorjsembe2024conditional,lee2023improving} highlight substantial challenges to utilize 3D representation in volumetric translation model for their increased demands for computational resources and large datasets that are costly to acquire in medical imaging. To the best of our knowledge, within the domain of 3D medical image inverse problems, no fully 3D models have demonstrated superior accuracy over 2D-based models due to these practical limitations. Designing a 3D network of the same size with 2D models is suggested by~\cite{weight_inf}, which is a promising approach given the rich 3D context and strong pre-trained 2D models. However, this is generally infeasible with existing 2D diffusion models~\cite{saharia2022palette}, which require around 500 GB of GPU memory for batch size 1 training and an extremely long training time. As a result, current 3D diffusion models~\cite{dorjsembe2024conditional} are designed to be much smaller with insufficient capacity to demonstrate competitive performance in inverse problems. Recent advances in volume-to-volume translation have introduced methods that combine perpendicular 2D diffusion models~\cite{lee2023improving, averagingdiffusion}, achieving improved accuracy and volumetric consistency. However, these methods cannot model the distribution of the entire volume since the generated images are produced by an averaging of the 2D networks without 3D representations, resulting in limited realism in 3D.

To effectively introduce 3D representations into volumetric translation, we present Score-Fusion, a pioneering model for volumetric translation that directly and effectively captures the distribution of 3D volumes. Score-Fusion adopts a two-stage training strategy: (1) It first trains multiple 2D diffusion models~\cite{SR3} in perpendicular planes. (2) It then utilizes a 3D fusion network to produce the final translation in each diffusion step. Meanwhile, the Score-Fusion model is designed to start with a weighted average of 2D models following TPDM\cite{lee2023improving}, which reduces 3D training to a fine-tuning process. The hierarchical layers of the 3D model are also reformulated to learn the ensemble of 2D features, further enhancing training efficiency and performance. Additionally, by ensembling diffusion models conditioned on various input modalities, Score-Fusion seamlessly supports multi-modality fusion.

The mathematical intuition of Score-Fusion lies in the properties of diffusion models and their associated score functions~\cite{song_score}. As the score function models the gradient of the probability distribution, it is inherently suitable for an ensemble. Previous works~\cite{averagingdiffusion,improving2D} have also demonstrated this by showing strong performance with a straightforward weighted averaging of score functions. Consequently, Score-Fusion replaces the weighted averaging process with a 3D network, effectively incorporating 3D representations. To the best of our knowledge, Score-Fusion is the first work to perform diffusion model fusion in the score function space, which provides new insights for diffusion model ensembling. Score-Fusion can also function as a plug-and-play mechanism compatible with various combinations of 2D models from previous studies~\cite{improving2D, averagingdiffusion, li2024two}, consistently delivering performance improvements across various 2D backbones.

Score-Fusion has been evaluated in various MRI image processing tasks on the BraTS~\cite{baid2021rsna} and HCP~\cite{HCP} dataset, including image super-resolution and modality translation. Our experimental results demonstrate that Score-Fusion performs superior volume translation over current state-of-the-art (SoTA) models in accuracy, realism, and downstream task performance. By learning to ensemble perpendicular 2D models conditioned on different input modalities, Score-Fusion shows strong performance without retraining new 2D models.

\section{Related Work}
{\color{gray}
\label{sec:Related Work}

}

\textbf{3D medical image generation and translation.} Attempts have been made to generate dense 3D volumes for medical imaging. Direct 3D-based diffusion models \cite{dorjsembe2024conditional,weight_inf} face difficulties due to high computational and dataset demands, resulting in moderate accuracy in tasks like super-resolution. Patch-wise, slice-wise, or cascaded generation strategies have been utilized to accommodate high-dimensional data\cite{highcap_3D}. However, in such models, initial inaccuracies in the low-resolution base are propagated during the refinement stages and the patch-based refinement often struggles with maintaining global consistency across the image. 
Latent 3D models \cite{dorjsembe2024conditional, makeavolum, khader2023denoising} have been exploited to compress the 3D data into a low-dimensional latent space and train diffusion models with this compressed latent space. However, the process of reducing dimensionality also has high computational and dataset demands and can lead to substantial reconstruction errors. Sequential slice generation from autoregressive models~\cite{peng2023generating,zhu2024generative} or simultaneous multiple slice generation may mitigate this issue of error accumulation over slices. Yet, these approaches suffer from challenges in maintaining coherence for long-range structures. More related to our approach, integrating multiple 2D models trained along perpendicular directions is a promising approach. TPDM \cite{lee2023improving} first proposes to combine two perpendicular 2D diffusion models to improve 3D imaging, where the weighted average of scores from pre-trained 2D models estimates the score function of a 3D model. Building on this concept,TOSM~\cite{li2024two} and MADM~\cite{averagingdiffusion} further improve the model performance by including 2D models in all three directions and using multiple consecutive 2D slices in 2D models. These models generate highly accurate results by effectively leveraging the high-resolution information in each 2D plane. 

\begin{figure*}[t] \centering 
\vspace{0.2cm}
\includegraphics[trim={0 1cm 0 1cm}, width=1\textwidth]{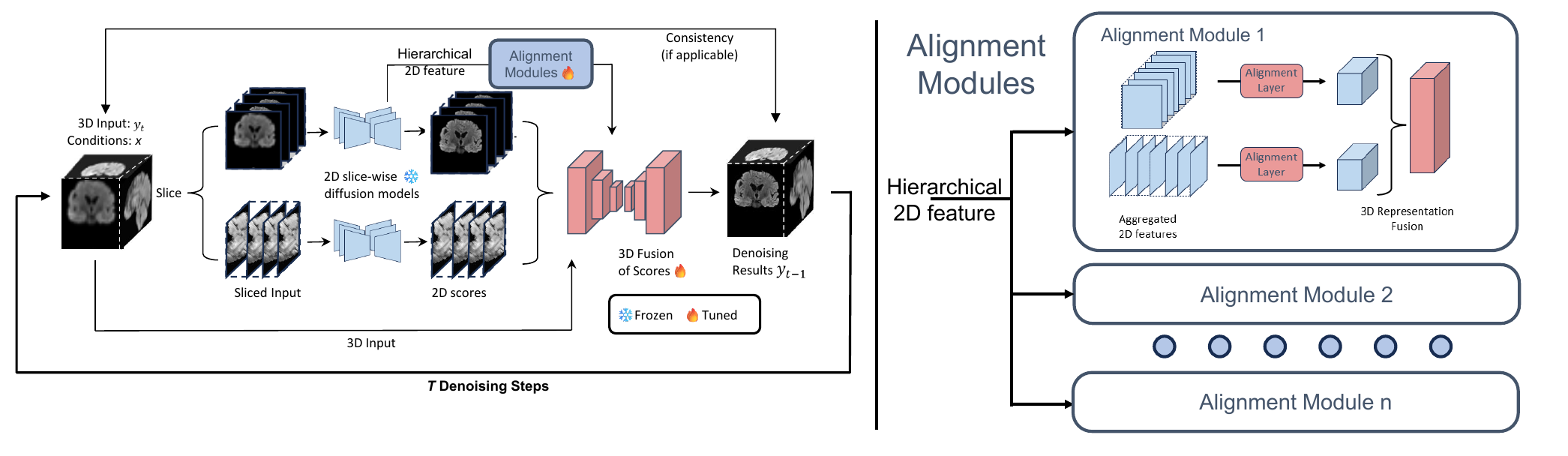} 
\caption{Overview of the Score-Fusion. At each denoising step, two pre-trained 2D models provide initial estimations of the scores in a slice-wise manner. Subsequently, a 3D network learns to integrate these estimations via 3D representation extracted from 3D input and aggregated 2D scores. In addition, the 3D model is initialized to output an average of 2D scores. Moreover, The 3D network layers are also reformulated to learn an ensemble of aggregated and aligned 2D features. These designs accelerate and stabilize the 3D training process.}
\label{fig:overview}
\vspace{-0.5cm}
\end{figure*}
\textbf{Model ensembling.}
Ensemble techniques, which include bagging, boosting, and stacking, have been further developed through specialized algorithms like Random Forest, AdaBoost, XGBoost, and Mixture of Experts (MoE). These techniques also demonstrated remarkable efficacy in medical image analysis, particularly in brain tumor segmentation \cite{SWINUnet, med_ensemble}, hypertension detection \cite{fitriyani2019development}, and kidney stone identification \cite{kazemi2018novel}. More recent research underscores the potential of ensembles as an effective strategy for scaling up large models\cite{moe}. 

\textbf{Diffusion model ensembling.} Recently, diffusion models have shown great success. The ensembling methods for diffusion models have become a useful research topic. Most current works use weighted averages to ensemble different branches of diffusion models~\cite{ssdiff, lee2023improving}. Collaborative Diffusion~\cite{cola-diff-face} is a learning-based ensembling method for diffusion; it trains an auxiliary model to estimate the confidence score for each branch of diffusion and ensemble based on the score. In this work, our approach uses information across all branches of diffusion models. This under-explored approach provides new insights for advancing diffusion model ensembling.

\section{Score Fusion in 3D}

\textbf{Problem formulation.}
We formulate volume-to-volume translation as a conditional sampling problem. Specifically, let 
\(\mathbf{x} \in \mathbb{R}^{b_1 \times b_2 \times b_3}\)\ be an input medical image volume, and let 
\(\mathbf{y} \in \mathbb{R}^{b_1 \times b_2 \times b_3}\)\ be the corresponding target volume to be generated, where \(b_1\), \(b_2\), and \(b_3\) denote the spatial dimensions of the volume. Our objective is to learn a conditional distribution 
\(p(\mathbf{y} \mid \mathbf{x})\) that accurately captures the volumetric structure in three dimensions. The input volume \(\mathbf{x}\) may consist of low-resolution data and/or a different imaging modality.

\subsection{Overall Framework of Score-Fusion}
\label{sec:pre}
\label{sec:main_method}
We designed the Score-Fusion as a conditional diffusion model. Following DDPM and Palette~\cite{DDPM,saharia2022palette}, our model gradually adds Gaussian noise to the target image in the training dataset during the \textit{forward} or \textit{diffusion process} as: 
$
q\left(\bm{y}_{t} | \bm{y}_{t-1}\right)= \mathcal{N}\left(\bm{y}_{t} ; \sqrt{1-\beta_{t}} \bm{y}_{t-1}, \beta_{t} \bm{I}\right); \quad  q(\bm{y}_{T}|\bm{y}_{0}) = q(\bm{y}_0) \prod_{t=1}^{T} q(\bm{y}_t | \bm{y}_{t-1})$, where $\bm{y}_0 \sim q(\bm{y})$ is the target image and $\beta_t$ is the variance of noise added at timestep $t$. The forward process produces a sequence of increasingly noisy variables $\bm{y}_1,..., \bm{y}_T$, after sufficient noising steps, the process reaches a pure Gaussian noise, \ie, $\bm{y}_T \sim \mathcal{N}(\bm{0},\bm{I})$. 

During \textbf{training}, our denoising diffusion model, $\bm{\epsilon}_{\theta}(\bm{y}_{t}, \bm{x}, t)$, is trained to predict the noise added into $\bm{y}$, given $\bm{y}_t$. Demonstrated effective in existing work \cite{SR3}, the sampling process can be guided by concatenating the noisy image $\bm{y}_t$ with condition $\bm{x}$. The loss used to optimize $\bm{\epsilon}_{\theta}(\bm{y}_{t}, \bm{x}, t)$ is: $\big\|\bm{\epsilon}_{\theta}(\sqrt{\alpha_{t}} \bm{y}_0+\sqrt{1-\alpha_{t}}\bm{\epsilon}, \bm{x}, t) - \bm{\epsilon}\big\|_2^2 $, where $\alpha_{t}:=\prod_{i=1}^{t}\left(1-\beta_{i}\right)$, and we sample $\bm{y}_0, \bm{x} \sim p(\bm{y}_0, \bm{x})$,  $\epsilon \sim \mathcal{N}(\bm{0},\bm{I})$.

During \textbf{sampling} in the \textit{reverse} or \textit{generative process}, we also follow Palette \cite{saharia2022palette} to generate images by iteratively removing the added noise in the sequence $\bm{y}_{T-1}, ..., \bm{y}_1, \bm{y}_0$, from a standard Gaussian prior $\bm{y}_T \sim \mathcal{N}(\bm{0},\bm{I})$. In addition, inspired by \cite{scoreSDE, DPS, song2023solving}, we explore self-consistency for solving inverse problems. More specifically, in each diffusion sampling step, we estimate noise with our denoising model $\bm{\epsilon}_{\theta}(\bm{y}_t, \bm{x},t)$. Therefore, we have the estimated $\hat{\bm{y}}_0(t)$ at the t-th denoising step as $\hat{\bm{y}}_0(t) := (\bm{y}_t - \sqrt{1-\alpha_{t}}\bm{\epsilon}_{\theta}(\bm{y}_t, \bm{x},t))/\sqrt{\alpha_{t}}$. In inverse problems, conditional input $\bm{x}$ is obtained through a known linear degradation process $\bm{x} = A\bm{y}$. At each diffusion step, we project the estimated $\bm{\hat{y}}_0(t)$ to a plausible $\bm{\hat{y}}_0(t)$, such that $\bm{x} = A\bm{\hat{y}_0}$, by $\bm{\hat{y}}_0(t) \gets \bm{\hat{y}}_0(t) - A^T{(AA^T)}^{-1}(A\bm{\hat{y}}_0(t) - x)$. We provide more details in Sec.~\ref{sec:consistency}. After the consistency projection, we obtain $\bm{y}_{t-1}$ by adding noise back: $\bm{y}_{t-1} = \sqrt{\alpha_{t-1}} \bm{\hat{y}}_0(t) + \sqrt{1-\alpha_{t-1}}\bm{\epsilon}$.

\textbf{Score-Fusion Models.} The key component of this work lies in our denoising network $\bm{\epsilon}_{\theta}(\bm{y}_{t}, \bm{x}, t)$. In particular, our model $\bm{\epsilon}_{\theta}$ consists of two 2D diffusion denoising models, $\bm{\epsilon}_{\theta a}^{2D(a)} $ and $ \bm{\epsilon}_{\theta b}^{2D(b)} $,  and a 3D diffusion denoising model, $ \bm{\epsilon}_{\theta _{3D}}^{3D} $, with $\theta_a$, $\theta_b$, and $\theta_{3D}$ being their trainable parameters, respectively. The 3D network is conditioned on two 2D diffusion models, trained to capture 2D image distributions along orthogonal planes to provide complementary views of the volumetric data. The 3D network is carefully initialized to start with a weighted average of 2D networks' estimation, such that Score-Fusion has the same performance with TPDM\cite{lee2023improving} before any 3D training. This design effectively constrains the 3D model, reduces the 3D training to a fine-tuning process, and thus promotes faster and stabilized training convergence. Additionally, the hierarchical representations from the 2D models are introduced to the layers in the 3D model with alignment projection layers. In this way, the 3D model's layers are reformulated to learn an ensemble of pre-trained 2D models' representations, instead of learning representations from scratch. Therefore, the training of the 3D model is further accelerated and stabilized by using the aligned 2D representation as a reference. We refer to this hybrid 2D/3D volumetric generative model as Score-Fusion. Fig.~\ref{fig:overview} provides a schematic overview of Score-Fusion.

\subsection{2D Score Models}
The 2D diffusion models, $ \bm{\epsilon}_{\theta a}^{2D(a)} $ and $ \bm{\epsilon}_{\theta b}^{2D(b)} $,  are trained on two perpendicular slices of the volumes using a standard conditional diffusion framework \cite{SR3}. 
 We take gradient descent steps on the following objectives for both 2D diffusion models during training:
\begin{equation}
      \begin{aligned}
     & \nabla_{\theta a} \big\|\bm{\epsilon}^{2D(a)}_{\theta a}(\bm{y}_t[:,i,:] , \bm{x}[:,i,:], t) - \bm{\epsilon}\big\|_2^2 \\
      & \nabla_{\theta b} \big\|\bm{\epsilon}^{2D(b)}_{\theta b}({\bm{y}_t[:,:,j]} , \bm{x}[:,:,j], t) - \bm{\epsilon}\big\|_2^2 
      \end{aligned}
      \vspace{-0.2cm}
\end{equation}

Here, $i$ and $j$ are the indices for the slices along two perpendicular planes, which are sampled uniformly: $ i \sim \text{Uniform}\{0,...,b_2\}$, $ j \sim \text{Uniform}\{0,...,b_3\} $. After proper training, the high-capacity 2D model can provide a decently accurate estimation of $\bm{\epsilon}$ for every volume slice.

\subsection{3D Fusion Model} 
\label{sec:3densembling}
The 3D ensembling model, $ \bm{\epsilon}_{\theta_{3D}}^{3D} $, is trained to fuse the pre-trained 2D diffusion models to capture the desired volumetric image distributions. In this stage, we first obtain the inference results, $  \hat{\bm{Y}}^{2D(a)} $ and $  \hat{\bm{Y}}^{2D(b)} $, from the 2D diffusion models, $ \bm{\epsilon}_{\theta a}^{2D(a)} $ and $ \bm{\epsilon}_{\theta b}^{2D(b)} $ by iterating through the sliced directions: 
\begin{equation}
\label{eq:train_2D_a}
\begin{aligned}
     & \hat{\bm{Y}}^{2D(a)}[:,i,:] = \bm{\epsilon}^{2D(a)}_{\theta a}(\bm{y}_t[:,i,:], \bm{x}[:,i,:], t) 
     \text{  for } i \in [0,b_2) \\
     & \hat{\bm{Y}}^{2D(b)}[:,:,j] = \bm{\epsilon}^{2D(b)}_{\theta b}(\bm{y}_t[:,:,j], \bm{x}[:,:,j], t) 
     \text{  for } j \in [0,b_3)
     \end{aligned}
\end{equation}
During training of the 3D diffusion model, both 2D models return $\hat{\bm{Y}}$'s, which contains the predicted score and a hierarchical feature map of the model: $\hat{\bm{Y}}^{2D(a)} = (\hat{\epsilon}^{2D(a)}, \mathcal{F}^{2D(a)} ) $, $\hat{\bm{Y}}^{2D(b)} = (\hat{\epsilon}^{2D(b)}, \mathcal{F}^{2D(b)} ) $.

The 3D model is designed to learn an ensemble of multiple 2D models'  score estimation with 3D representation. Specifically, at each diffusion step, the 3D model takes as input the original image $\bm{x}$, the noisy intermediate state $\bm{y}_t$, and the aggregated score estimation $\hat{\epsilon}$ obtained from the 2D diffusion models. Furthermore, aggregated feature maps $\mathcal{F}$ from the 2D models are incorporated as supplementary information as in Fig.~\ref{fig:overview}. Formally speaking, the 3D network is trained to perform the ensembling process using the following formulation using an L2 loss: 

\begin{equation}
\label{eq:training_3D}
      \nabla_{\theta_{3D}} \big\|\bm{\epsilon}^{3D}_{\theta_{3D}}(\bm{y}_t, \bm{x}, \hat{\bm{Y}}^{2D(a)}, \hat{\bm{Y}}^{2D(b)}, t) - \bm{\epsilon}\big\|_2^2
\end{equation}

where $\bm{y}_0,$ $\bm{x} \sim p(\bm{y}_0, \bm{x})$, and $\epsilon \sim \mathcal{N}(\bm{0},\bm{I})$. Although, the inference results from 2D models, $\hat{\bm{Y}}^{2D(a)}$ and $\hat{\bm{Y}}^{2D(b)}$, already help the training of the 3D model, 3D ensemble model still needs to be trained from scratch. To improve training speed, we initially pre-train the model on 3D patches, $(\bm{y}_0, \bm{x}) = \text{crop}(\bm{y}_0, \bm{x})$, and then fine-tune it on the full volumes. Due to the translation invariance of our convolution-based networks, we empirically find that a naively pre-train on the patches results in a decently good network initialization, thereby effectively improving the training convergence. While existing works, such as~\cite{patch_diff}, could potentially enhance this patch-wise diffusion training process, we leave such optimizations for future work.

In this work, the \textbf{network architecture} of the 3D model, $\bm{\epsilon}^{3D}_{\theta_{3D}}$, is a 3D Unet-like denoising model with time-step embeddings, the 3D input $\bm{x}$, the noisy target $\bm{y}_t$, and the noise estimated by the 2D models $ \hat{\epsilon}^{2D(a)}, \hat{\epsilon}^{2D(b)} $ as the input of the 3D model. In the encoder, each downsampling block is enriched with corresponding feature maps from the features of both 2D models, $\mathcal{F}^{2D(a)}$, and $\mathcal{F}^{2D(b)}$. At each hierarchical level, the feature maps are first processed with MLP-based alignment layers, which align the 2D features with the 3D model and map them to an appropriate shape. The feature maps are then injected into 3D layers, providing the 3D layer with a reference to fused 2D features. Hence, the 3D layers are reformulated to learn an ensemble of the aggregated 2D features, which is easier than learning representation from scratch. In addition, using the feature maps mitigates the risk of information bottlenecks between the 2D and 3D stages, which could otherwise limit the performance. Additionally, rather than directly outputting the predicted noise, our 3D U-Net-like model produces two components: a weight vector $\bm{w}$, used to ensemble the estimations from the 2D models, and a residual term $R$, which is directly estimated by the 3D model. These two outputs are combined to form the final prediction:
\begin{equation}
    \bm{\epsilon}^{3D}_{\theta_{3D}}(...) = (0.5 + \bm{w})\hat{\epsilon}^{2D(a)} + (0.5 - \bm{w})\hat{\epsilon}^{2D(b)} + \lambda R
\end{equation}
where $\lambda$ is a hyperparameter, whereas $\bm{w}$ and $R$ are of the same size as the target noise $\epsilon \in \mathbb{R}^{b_1,b_2,b_3}$. This design enables the model to dynamically select the more reliable 2D estimation based on 3D context and allows the 3D model to contribute 3D-specific content $R$. Meanwhile, a tunable weight parameter, $\lambda$, controls the model’s reliance on the 3D output, $R$. In addition, inspired by ControlNet~\cite{controlnet}, a zero-initialized convolution layer at the end of the model ensures smooth initialization, making the 3D training a fine-tuning process starting with an average weighting strategy, and thereby stabilizing the 3D model training. The pseudo-code for training and inference with Score-Fusion is provided in Algorithm~\ref{alg:training} and Algorithm~\ref{alg:inference}.

\subsection{Multi-modality Fusion} 
\label{sec:details}

In volumetric translation for medical imaging, the conditions for translating a new image can be multifaceted. For instance, DDMM-Synth~\cite{li2023ddmm} suggested using both MRI and low-resolution CT scans to produce high-resolution CT images. Training a separate model for each possible combination of input conditions would result in exponential time complexity, making it generally impractical. Therefore, a model that can efficiently integrate pre-trained models across diverse conditions provides significant advantages. Score-Fusion achieves this by naturally integrating multiple diffusion models, each conditioned on individual modalities, through a 3D network architecture that functions similarly to fusing two 2D models described in~\ref{sec:3densembling}. This approach leverages the accelerated 3D training of Score-Fusion. To further enhance the speed of multi-modality fusion, we employ a smaller variant of our model, adjusting the number of channels in each layer.

\algrenewcommand\algorithmicindent{2em}%

\begin{algorithm}[h]
  \caption{Training of Score Fusion} \label{alg:training}
  \small
  \begin{algorithmic}[1] 
      \Repeat
    \State $(\mathbf{x}, \mathbf{y_0}) \sim p(\mathbf{x}, \mathbf{y_0})$ \Comment{sample from dataset}
    \If{pretrain}  \Comment{pretrain on patch}
    \State $(\mathbf{x}, \mathbf{y_0}) = \text{crop}(\mathbf{x}, \mathbf{y_0})$
    \EndIf
    
    \State $ t \sim \text{Uniform}(0,T)$ ; $ \epsilon \sim \mathcal{N}(\mathbf{0},\mathbf{I})$ 
   \For{$i = 0$ to $b_2$}
    \State $ \hat{\mathbf{Y}}^{2D(a)}[:,i,:] \gets \bm{\epsilon}^{2D(a)}_{\theta a}(\mathbf{y}_t[:,i,:], \mathbf{x}[:,i,:], t)$
    \EndFor
    \For{$j = 0$ to $b_3$}
    \State $ \hat{\mathbf{Y}}^{2D(b)}[:,:,j] \gets \bm{\epsilon}^{2D(b)}_{\theta a}(\mathbf{y}_t[:,:,j], \mathbf{x}[:,:,j], t)$
    \EndFor
    \State Take a gradient descent step on 
    \State $\nabla_{\theta_{3D}} \left\|\bm{\epsilon}^{3D}_{\theta_{3D}}(\mathbf{y}_t, \mathbf{x}, \hat{\mathbf{Y}}^{2D(a)}, \hat{\mathbf{Y}}^{2D(b)}, t) - \epsilon \right\|_2^2$
    \Until{converged}
    
  \end{algorithmic}
 
\end{algorithm}

\begin{algorithm}
  \caption{Inference of Score Fusion} \label{alg:inference}
  \small
  \begin{algorithmic}[1]
    
    \State $(\bm{x}) \sim p(\bm{x})$ \Comment{sample from dataset}
        
    \State $ \bm{y_T} \sim N(0,1)$
    \For{t = T,...,1,0}
        \For{i = 0,1,...,$b_2$}
            \State $ \hat{\bm{Y}}^{2D(a)}[:,i,:] \gets \bm{\epsilon}^{2D(a)}_{\theta a}(\bm{y}_t[:,i,:], \bm{x}[:,i,:], t)$
        \EndFor
        \For{j = 0,1,...,$b_3$}
            \State $ \hat{\bm{Y}}^{2D(b)}[:,:,j] \gets \bm{\epsilon}^{2D(b)}_{\theta a}(\bm{y}_t[:,:,j], \bm{x}[:,:,j], t) $ 
        \EndFor
        \State $\hat{\bm{\epsilon}}^{3D} = \bm{\epsilon}^{3D}_{\theta}(\bm{y}_t, \bm{x},\hat{\bm{Y}}^{2D(a)}, \hat{\bm{Y}}^{2D(b)} , t)  $ 
        \State $\bm{\hat{y}}_0 \gets \frac{\bm{y}_t - \sqrt{1-\alpha_{t}}\bm{\epsilon}^{3D}}{\sqrt{\alpha_{t}}}$ \Comment{get current estimation of y0}
        \If{ Inverse problem } 
            \State $\bm{\hat{y}}_0 \gets \bm{\hat{y}}_0 - A^T{(AA^T)}^{-1}(A\bm{\hat{y}}_0 - x)$
        \EndIf
        \State $\bm{y}_{t-1} \gets \sqrt{\alpha_{t-1}} \bm{\hat{y}}_0 + \sqrt{1-\alpha_{t-1}}\bm{\epsilon}$ 
        
    \EndFor
    \State return $\bm{y}_0$
  \end{algorithmic}
\end{algorithm}

\section{Experiments}

\subsection{Experimental Setup}
\label{exp:experimentalsetup}
\textbf{Datasets.} We conducted experiments using the BraTS 2021 training dataset \cite{baid2021rsna}, which includes 1,251 volumetric brain scans with tumors across 4 modalities: FLAIR, T1, T1ce, and T2. We randomly divided the dataset into a 0.8:0.2 split for training and evaluation purposes, allowing its use for downstream tasks as well. Each scan was center-cropped to a dimension of 192$\times$192$\times$152 to remove the blank background. For training 2D models, we sliced the 3D volumes in two directions---transverse and sagittal planes for both TPDM baselines and Score-Fusion. In the super-resolution experiment, FLAIR images were downsampled using [4$\times$4$\times$4] average pooling. For modality translation, T1ce images served as inputs with FLAIR images as targets. Additionally, we investigated a multi-condition task, using both low-resolution FLAIR and T1ce images as input to predict high-resolution FLAIR images. In addition, we investigate Score-Fusion’s generalizability to different datasets by applying it, alongside related baselines, to a super-resolution task on the FLAIR modality of the HCP dataset \cite{HCP}. This demonstrates Score-Fusion’s potential for broader applicability. We present both quantitative and qualitative results for the HCP dataset in Section~\ref{sec:HCP}.

\textbf{Baselines.} 
We reproduced diverse baseline methods across a diverse set of established 2D and 3D translation models to ensure a comprehensive comparison. For slice-wise 2D models, we utilized Pix2pix~\cite{isola2017image} as the representative GAN-based method, U-Net~\cite{unet} for supervised regression, Palette~\cite{saharia2022palette} as a diffusion-based approach, and I2SB~\cite{liu20232} for optimal-translation-based modeling. Similarly, for 3D-based baselines, we used Pix2pix3D~\cite{isola2017image}, U-Net3D~\cite{unet}, Med-DDPM~\cite{dorjsembe2024conditional} (or Palette3D). As stated in Sec.~\ref{sec:Introduction},  Med-DDPM uses a small denoising network and thus demonstrates limited performance. In addition, we used Palette-2.5D for another baseline, which uses multiple consecutive 2D slices as input.
Several existing approaches closely related to our method combine multiple pre-trained 2D diffusion models in perpendicular orientations, demonstrating enhanced performance over other baselines. For instance, TPDM~\cite{lee2023improving} combines two 2D diffusion models trained on perpendicular planes. To support modality translation, we adapted the TPDM's 2D backbone to Palette~\cite{saharia2022palette} architecture in place for DPS~\cite{DPS}. Furthermore, TOSM~\cite{li2024two} employs three perpendicularly trained 2D diffusion models, whereas MADM~\cite{averagingdiffusion} uses three 2.5D diffusion models. We perform a hyper-parameter search on the super-resolution task on the BraTS dataset for all baselines. 

\textbf{Model Architecture and Variants.} 
To make a fair comparison, we use pre-trained 2D models from TPDM, TOSM, and MADM utilizing an existing 3D diffusion model architecture, Med-DDPM~\cite{dorjsembe2024conditional}. The TPDM-based model is our primary model as it achieves 30\% faster inference speed and smaller model size relative to the TOSM-based model, as shown in Tab.~\ref{tab:speed}. Meanwhile, MADM-based and TOSM-based models are heavier variants that yield performance gains across all metrics. These consistent improvements in all three variants demonstrate that our approach can serve as a plug-in-and-play mechanism for multiple combinations of 2D/2.5D model backbones and existing 3D model architectures. We also include a more detailed model architecture in Sec.~\ref{sec:architecture}.

\textbf{Metrics.} 
We used multiple metrics to assess both the accuracy and realism of generated MRI images. For accuracy, we used peak signal-to-noise ratio (PSNR) and the structural similarity index measure (SSIM), which are widely used in medical imaging. To evaluate perceptual quality and realism, we used the maximum mean discrepancy (MMD)~\cite{gretton2012kernel} and the Fréchet inception distance (FID) metrics~\cite{heusel2017gans}. 
Lower MMD/FID scores imply the generated images are more realistic. To evaluate the FID score, following common practice~\cite{dorjsembe2024conditional, sun2022hierarchical}, we adopted the same pre-trained model~\cite{chen2019med3d} to extract features and calculate the FID metrics in the feature space. Because diffusion-based models exhibit inherent stochasticity, we further assess uncertainty by performing inference multiple times with different noise realizations $\epsilon$. From these runs, we calculate voxel-wise means and standard deviations, thereby providing uncertainty-aware metrics. Quantitative and qualitative results for these metrics can be found in Sec.~\ref{sec:Uncertainty}.

\begin{figure*}[h] \centering 
\includegraphics[width=0.8\textwidth]{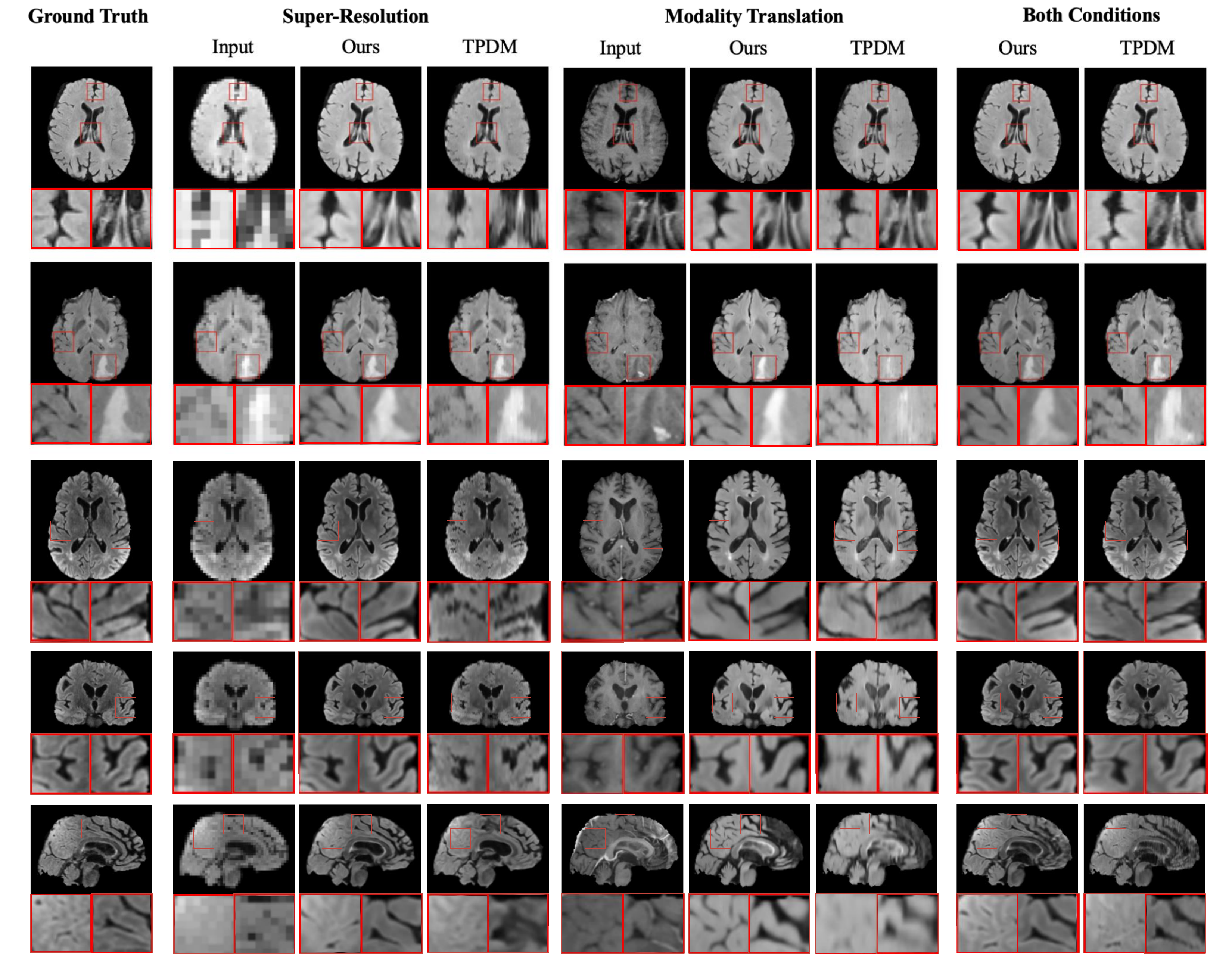} 
\vspace{-0.5cm}
\caption{Visual comparison of generated samples for three different conditions. The first three rows show axial view slices from different MRI volumes. Neither Score-Fusion nor TPDM have a 2D model trained in this direction. The last three rows show slices for the same MRI volume in all three views. Score-Fusion reconstructs more realistic details with smoother edges and fewer artifacts.}
\label{fig:zoomin}
\vspace{-0.3cm}
\end{figure*}

\subsection{Experimental Results}

We showcase the performance of Score-Fusion in solving various translation problems, including 4$\times$ super-resolution, modality translation, and a fusion of both conditions in Fig.~\ref{fig:zoomin}. We also include more randomly selected samples for more variants of our model in Sec.~\ref{sec:visualization}. Fig.~\ref{fig:zoomin} shows the generation quality under various conditions and provides comparisons with other methods. The first two columns show the performance for super-resolution and modality translation, and the last columns show the model performance when fusing these two conditions. Our approach excels in faithfully recovering intricate high-frequency details, particularly in tumor-affected areas where such details are complex and often underrepresented. In the super-resolution task, from the zoomed-in panel, Score-Fusion clearly distinguishes tissue boundaries across various tissue types, including tumor and white/grey matter. In the modality translation task, the distribution of contrast difference between modalities is also better captured, as stated in Sec.~\ref{sec:Uncertainty}. Furthermore, our method demonstrates superior volumetric consistency, while the baseline model exhibits noticeable artifacts. In the visualization of Score-Fusion, the fidelity of tissue texture and sharpness along all three orthogonal directions are well preserved, even though the 2D diffusion models in Score-Fusion are trained only on transverse and sagittal planes. Our model reconstructs tumor regions with clearer margins, fewer artifacts, and higher resolution for samples containing tumors, providing superior performance in all three orthogonal directions. Overall, Score-Fusion generates images with higher accuracy, realism, and volumetric consistency. Tab.~\ref{tab:metrics} summarizes the quantitative results of each translation task. Score-Fusion clearly surpasses other baseline models in most metrics, showing superior fidelity, structure, texture preservation, and noise suppression performance. 

\begin{table*}
\vspace*{2mm}
\setlength{\tabcolsep}{3.5pt}
\small\centering
\scalebox{0.9}
{
\begin{tabular}{lcccccccccccc}
\toprule    
 \multirow{2.5}{*}{Method} & \multicolumn{4}{c}{SR} & \multicolumn{4}{c}{MT} & \multicolumn{4}{c}{both condition}  \\
 \cmidrule(lr){2-5}
 \cmidrule(lr){6-9}
 \cmidrule(lr){10-13}
 & PSNR($\uparrow$) & SSIM($\uparrow$) & MMD($\downarrow$) & FID(1e-4)($\downarrow$) & PSNR & SSIM & MMD & FID & PSNR & SSIM & MMD & FID \\
\midrule
Pix2pix\cite{isola2017image} & 28.75 & 0.889 & 512.2 & 25.9 & 22.25 & 0.812 & 8989.0 & 577.6 & 31.78 & 0.923 & 133.9 & 11.8 \\
U-net\cite{unet} & 30.32 & 0.579 & 917.2 & 58.9 & 23.74 & 0.846 & 1829.0 & 320.3 & 33.58 & 0.931 & 83.5 & 36.6 \\
Palette\cite{saharia2022palette} & 29.26 & 0.894 & 40.9 & 13.5 & 22.68 & 0.784 & 284.4 & 85.9 & 33.6 & 0.939 & 34.9 & 9.3\\
I2SB\cite{liu20232}  & 27.51 & 0.860 & 2644.5 & 47.7 & 20.75 & 0.738 & 35774.5 & 1343.6 & 31.3 & 0.905 & 1313.6 & 12.0 \\
\midrule
Palette-3D\cite{dorjsembe2024conditional}  & 28.48 & 0.320 & 4222 & 88.7 &\multicolumn{4}{c}{Not Working} & 24.98 & 0.297 & 15926.0 & 463.1 \\
Pix2pix-3D\cite{isola2017image} & 29.54 & 0.866 & 516.5 & 8.6 & 22.77 & 0.784 & 1974.0 & 342.2 & 31.86 & 0.900 & 87.81 & 56.2 \\
U-net-3D\cite{unet} & 31.23 & 0.892 & 115.6 & 59.6 & 23.43 & 0.809 & 487.5 & 273.8 & 32.95 & 0.922 & 43.3 & 43.9 \\
Palette-2.5D\cite{saharia2022palette}  & 29.76 & 0.834 & 35.37 & 12.3 & 23.04 & 0.728 & 1141.19 & 258.6 & 25.89 & 0.819 & 2858.37 & 138.92\\

\midrule
TPDM\cite{lee2023improving}  & 32.23 & 0.922 & 29.35 & 17.3 & 25.35 & 0.868 & 176.3 & 185.5 & 35.12 & 0.945 & 14.5 & 22.2 \\
Ours-TPDM & 33.24  & 0.944 & 13.77 & 8.31 & 25.26 & 0.882 & 154.9 & \textbf{48.2} & 36.24 & 0.961 & 7.52 & 5.8 \\
\midrule
TOSM\cite{li2024two}  & 32.76 & 0.932 & 24.17 & 24.87 & \textbf{25.66} & 0.881 & 1018.91 & 209.5 & 35.44 & 0.947 & 8.32 & 14.53 \\

Ours-TOSM & 33.30 & \textbf{0.945} & 13.62 & \textbf{6.51} & 25.24 & \textbf{0.882} & \textbf{138.47} & 136.06 & \textbf{36.51} & 0.963 & 5.926 & \textbf{3.72} \\

MADM\cite{averagingdiffusion}  & 33.02 & \textbf{0.945} & 30.92 & 35.64 & 25.47 & 0.874 & 1419.4 & 251.4 & 35.21 & 0.946 & 8.21 & 13.6 \\

Ours-MADM & \textbf{33.31} & \textbf{0.945} & \textbf{13.46} & 6.57 & 25.13 & 0.876 & 192.51 & 130.33 & 36.37 & \textbf{0.964} & \textbf{5.44} & 3.81 \\

\bottomrule

\end{tabular}
}
\caption{Quantitative evaluation of Score-Fusion on BraTS dataset. Best metrics are highlighted in \textbf{bold}. The proposed model achieves better accuracy (PSNR/SSIM) given more 3D context than their corresponding variant in most tasks. Moreover, thanks to 3D representation, Score-Fusion achieves significantly better 3D realism (MMD/FID). We demonstrate the standard deviation and uncertainty metrics in Tab.~\ref{tab:uncertain}.}

\label{tab:metrics}
\vspace{-0.5cm}
\end{table*}

\subsection{Downstream Task}
\label{sec:downstream}
We further evaluated the performance of Score-Fusion on a downstream task: tumor segmentation, where high-quality input modalities are crucial for accurately segmenting complex structures. Using the BraTS 2021 tumor segmentation dataset, we applied a pre-trained SwinUNetR~\cite{SWINUnet} model with four modalities, replacing the ground-truth FLAIR modality with inferences from each model. Segmentation performance was assessed on Tumor Core (TC), Whole Tumor (WT), and Enhancing Tumor (ET) regions using two metrics: Dice score and Recovery rate. The Dice score measures segmentation quality, while the Recovery rate quantifies each model's segmentation score recovered from low-quality FLAIR to the generated FLAIR, with segmentation using ground truth (GT) FLAIR as the upper bound and downsampled FLAIR as the lower bound. The Recovery rate is defined as: \text{Recovery Rate} = (\text{Prediction} - \text{Downsample}) / (\text{Ground-Truth} - \text{Downsample}), where \textbf{Prediction} refers to the segmentation performance using predicted FLAIR, \textbf{Downsample} is the performance with downsampled FLAIR, and \textbf{Ground Truth} is the performance with ground-truth FLAIR.
As shown in Tab.~\ref{tab:downstream}, our methods consistently outperformed other methods. Our method also qualitatively results in smoother tumor edges and more accurate structures demonstrated in Fig.~\ref{fig:downstream_SR1}~\ref{fig:downstream_SR2}. (see Sec.~\ref{sec:downstream_detail} for more details).

\subsection{Multi-modality fusion}
\label{sec:multi_res}
As discussed in Sec.~\ref{sec:details}, the Score-Fusion not only merges 2D models trained in different directions but also effectively integrates models pre-trained under various single conditions when faced with new combinations of input modalities given pre-trained 2D models on every single condition. We show the model's performance in Tab.~\ref{tab:cross}. TPDM uses a weighted average for all 2D models, demonstrating limited performance. Using a 3D model of the original size, Score-Fusion learns to fuse scores estimated in two different conditions, demonstrating competitive performance without re-training 2D models. Score-Fusion-small further improves training speed with marginal performance drop to flexibly support multi-modality fusion. The models on both conditions (last 2 rows) show the metrics when re-training every 2D model on both conditions, representing an upper limit of multi-modality fusion performance. All training experiments are performed on Nvidia RTX A100 GPU.

\begin{table}
\setlength{\tabcolsep}{2pt}
\scriptsize\centering %
\scalebox{0.9}{
\vspace{-0.5cm}
\begin{tabular}{lcccccc}
\toprule
\multirow{2}{*}{Method} & \multicolumn{3}{c}{Dice (\%)} & \multicolumn{3}{c}{Recovery (\%)} \\
\cmidrule(lr){2-4} \cmidrule(lr){5-7}
                     & TC & WT & ET & TC & WT & ET \\
                     
\midrule
GT FLAIR             & 82.71 & 89.17 & 81.20 & -     & -     & -     \\
Downsampled GT       & 82.30 & 86.82 & 80.30 & -     & -     & -     \\
\midrule
SR                   &       &       &       &       &       &       \\
\quad TPDM           & 82.49 & 87.77 & 80.49 & 46.27 & 40.46 & 20.62 \\
\quad TOSM           & 82.52 & 87.21 & 80.80 & 54.55 & 16.51 & 55.28 \\
\quad ours-TPDM      & \textbf{82.69} & 87.85 & \textbf{80.94} & \textbf{93.71} & 43.80 & \textbf{71.69} \\
\quad ours-TOSM      & 82.59 & \textbf{87.86} & 80.87 & 70.14 & \textbf{44.38} & 63.94 \\
\midrule
MT                   &       &       &       &       &       &       \\
\quad TPDM           & 77.28 & 77.74 & 78.37 & -     & -     & -     \\
\quad TOSM           & 77.94 & \textbf{79.21} & 78.64 & -     & -     & -     \\
\quad ours-TPDM      & 77.88 & 78.51 & 78.22 & -     & -     & -     \\
\quad ours-TOSM      & \textbf{78.84} & 78.73 & \textbf{79.52} & -     & -     & -     \\
\midrule
both condition       &       &       &       &       &       &       \\
\quad TPDM           & 82.45 & 87.69 & 80.74 & 36.31 & 37.25 & 49.28 \\
\quad TOSM           & 82.54 & 87.27 & 80.82 & 57.81 & 19.25 & 57.79 \\
\quad ours-TPDM      & 82.46 & 87.91 & 80.74 & 38.66 & 46.67 & 48.97 \\
\quad ours-TOSM      & \textbf{82.61} & \textbf{87.98} & \textbf{80.89} & \textbf{75.35} & \textbf{49.69} & \textbf{65.72} \\
\bottomrule
\end{tabular}
}
\caption{Segmentation performance with the FLAIR modality replaced by model predictions.}
\label{tab:downstream}
\vspace{-0.3cm}
\end{table}

\subsection{Training and Inference Speed.} 
We show Training and Inference Speed in Tab.~\ref{tab:main_speed}. Previous 3D diffusion method struggles to use 3D representation, with extremely long training time (120 GPU days) and sub-optimal accuracy performance as in Tab.~\ref{tab:metrics}. In contrast, Score-Fusion effectively introduced 3D representation in just 16 days of extra 3D training time on top of TPDM. Score-Fusion-small further accelerates the 3D training for 4 times, achieving 30x more efficient than 3D diffusion baselines.

\begin{table}
\setlength{\tabcolsep}{4pt}
\small\centering
\vspace{-0.4cm}
\scalebox{0.8}{
\begin{tabular}{ccccc}
\toprule    
\multirow{2}{*}{Method} & \multirow{2}{*}{PSNR} & \multirow{2}{*}{SSIM} & \multirow{2}{*}{MMD} & {Training Time}\\
 &  &  &  & (GPU days) \\
\midrule
TPDM & 32.43 & 0.929 & 25.05 & 0 \\
Score-Fusion-small & 35.34 & 0.956 & 8.64 & 4 \\
Score-Fusion & 35.6 & 0.958 & 8.82 & 16 \\
TPDM-both\_cond & 35.12 & 0.945 & 14.5 & 8 \\
Score-Fusion-both\_cond & 36.24 & 0.961 & 7.52 & 24 \\

\bottomrule
\end{tabular}
}
\caption{Multi-modality fusion results for Score-Fusion. }
\label{tab:cross}
\vspace{-0.3cm}
\end{table}

\begin{table}[h]
\centering
\scalebox{0.8}{

\begin{tabular}{cccc}
\toprule
\multirow{2}{*}{\textbf{Time}} & \textbf{2D Training} & \textbf{3D Training } & \textbf{Inference } \\ 
& (GPU days) & (GPU days) & (minutes/volume) \\
\midrule

3D Palette & 0 & 120 & 0.6 \\ 
TPDM & 16 & 0 & 1.72 \\ 
Ours-TPDM & 16 & 16 & 2.34 \\ 
Ours-TPDM-small & 16 & 4 & 1.92 \\
\bottomrule
\end{tabular}

}
\caption{Training and Inference time results. GPUs are A100-40G. More complete comparison in Tab.~\ref{tab:speed}}
\label{tab:main_speed}
\vspace{-0.3cm}
\end{table}

\section{Conclusion}
3D voxel-space representation can be essential in medical image translation and generation for both volumetric realism and downstream task performance. However, existing models struggle to use 3D representation due to computational challenges and data scarcity. In this work, we have introduced Score-Fusion to effectively introduce 3D voxel space representation into 3D medical image translation by fusing estimations from slice-wise 2D models in the score function space. Several key designs, including average-initialization, feature map fusion, patch-wise pre-training. In addition, our model integrated the strengths of both 2D and 3D diffusion models. Score-Fusion provides strong insights for diffusion model ensembling as the first work to adopt a learning-based fusion in the score function space. Empirical evaluations on various 3D MRI image translation tasks, including super-resolution and modality translation, have shown that Score-Fusion achieves unmatched accuracy, volumetric realism, and downstream task performance. In addition to computational and memory efficiency, the approach offers considerable flexibility in merging models conditioned on different domains.

\section*{Impact Statement}
This paper presents work whose goal is to advance the field of Machine Learning and its application in medical imaging.
There are several \textbf{Limitations} of our work. Unlike some other multi-stage models~\cite{mapprior,cola-diff-face}, Score-Fusion struggles with joint end-to-end training due to the substantial computational demands of simultaneously managing high-capacity 2D models and the volumetric complexities of 3D tasks. In addition, the model's dependency on patchwise pre-training for efficient 3D model learning presents limitations for tasks requiring the integration of long-range spatial information, such as large-area inpainting and compressed sensing MRI. Therefore, Score-Fusion may require longer training for such tasks. There are also many potential societal consequences of our work. However, direct application of our method to medical imaging should be approached with caution.

\section*{Acknowledgement}
This work used the Delta system at the National Center for Supercomputing Applications through allocations CIS230243 and CIS240171 from the Advanced Cyberinfrastructure Coordination Ecosystem: Services \& Support (ACCESS) program, which is supported by National Science Foundation grants \#2138259, \#2138286, \#2138307, \#2137603, and \#2138296. We also thank Xiaoyue Li for the valuable discussion.

{
    \small
    \bibliography{main}
    \bibliographystyle{icml2025}
}

\newpage
\appendix
\onecolumn
\clearpage
\setcounter{page}{1}

\section{Overview}
In this supplementary material, we first discuss the uncertainty awareness results performed by our model and baselines by running the inference multiple times in Sec.~\ref{sec:Uncertainty}. We provide more randomly selected results (We do exclude samples with low-quality GT) for more baselines and our variants in Sec.~\ref{sec:visualization}. Then, we provide our super-resolution result in an additional dataset, HCP dataset~\cite{HCP} in Sec.~\ref{sec:HCP}. We provide ablation studies on key techniques in Sec.~\ref{sec:ablation}. We also provide more details on training and inference, including a detailed model architecture in Sec.~\ref{sec:architecture} and training/inference speed in Sec.~\ref{sec:speed}. We finally introduce a more detailed method for self-consistency projection in Sec.~\ref{sec:consistency} and downstream task results in Sec.~\ref{sec:downstream_detail}.

\section{Uncertainty Awareness}
\label{sec:Uncertainty}
As with most diffusion-based models, our models and some of our baselines can have uncertainty estimations. To study this uncertainty, we perform inference five times for each sample in our validation set. This gives us 5 PSNR and SSIM values for each data sample. We then calculate the standard deviation (std) of the PSNR and SSIM for each sample and include the mean std across the entire validation set in Tab.~\ref{tab:uncertain}. This further validates that our performance boost in PSNR and SSIM is significant. For the main variant, TPDM and Ours-TPDM, in the super-resolution task, we have a 1.01 boost in PSNR, which is much larger than the std of PSNR for both models (0.0066 and 0.0298). Even for MADM and Ours-MADM, where we have the most marginal PSNR boost, the boost is still 0.3, around 10 times larger than the std for both models (0.0308 and 0.0276). In contrast, in modality translation, the std is significantly larger since the uncertainty in this task is much larger than in others, indicating the PSNR drop is not as significant. In fact, previous work~\cite{SR3} argues that PSNR prefers blurry results, and highly diverse and realistic results typically have low PSNR in tasks with high uncertainty.

\begin{figure*}[h] \centering 
\includegraphics[width=\textwidth]{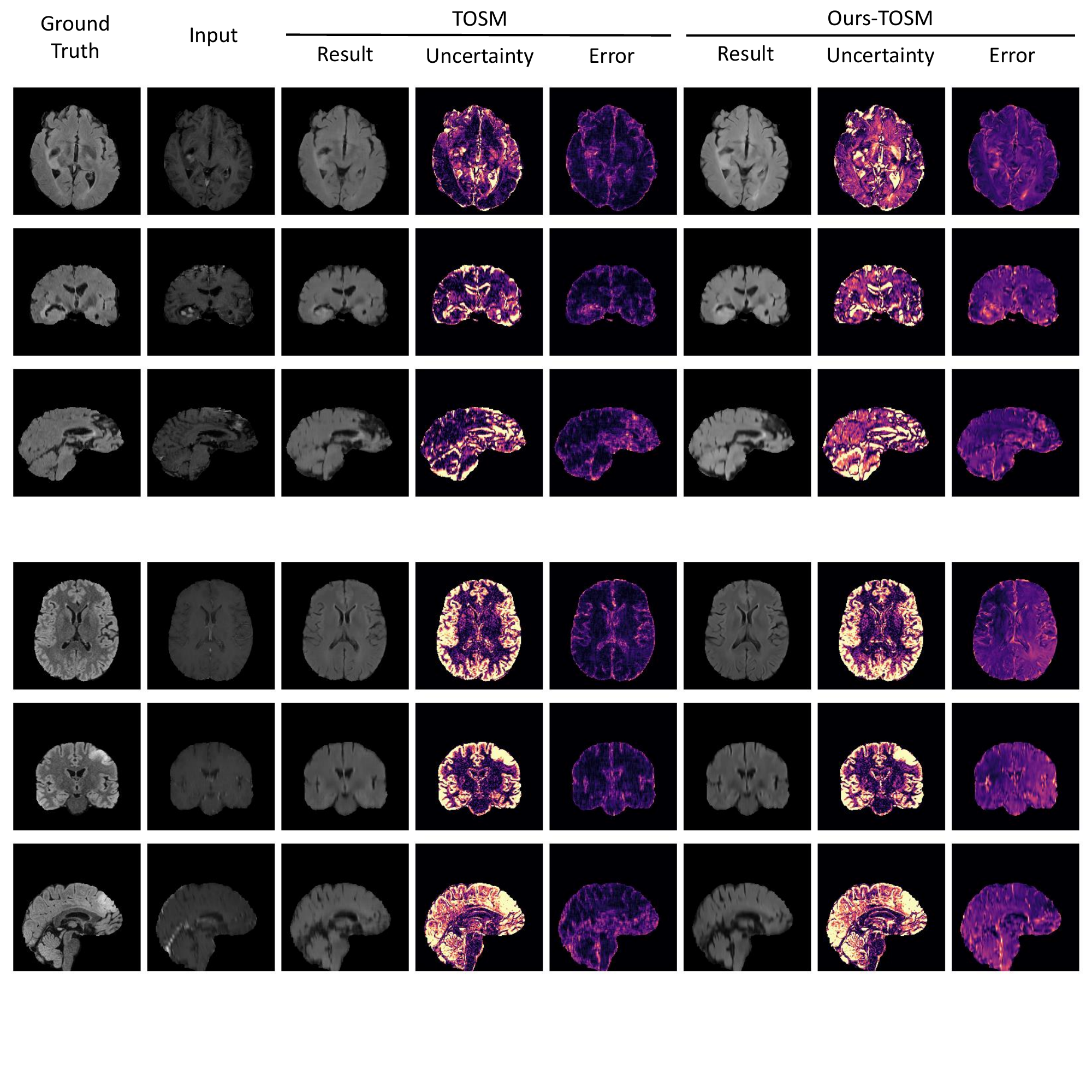} 
\vspace{-2cm}
\caption{Uncertainty awareness results on modality translation task}
\label{fig:MT2}
\end{figure*}

In addition, this inference also provides a mean $ \mu_i$ and std estimation $ \sigma_i$ for each voxel. We use Mean Absolute Calibration Error (MACE)~\cite{ece_regression} to measure the uncertainty awareness of our model and baseline. MACE measures the absolute difference between the predicted uncertainty and the actual error, as shown in Eq.~\ref{eq:MACE}.

\begin{equation}
\label{eq:MACE}
\text{MACE} = \frac{1}{N} \sum_{i=1}^{N} \left| \sigma_i - \left| y_i - \mu_i \right| \right|
\end{equation}
\begin{table*}
\vspace*{2mm}
\setlength{\tabcolsep}{3.5pt}
\small\centering
\scalebox{1.0}
{
\begin{tabular}{lccccccccc}
\toprule    
 \multirow{2.5}{*}{Method} & \multicolumn{3}{c}{SR} & \multicolumn{3}{c}{MT} & \multicolumn{3}{c}{both condition}  \\
 \cmidrule(lr){2-4}
 \cmidrule(lr){5-7}
 \cmidrule(lr){8-10}
 & PSNR($\uparrow$) & SSIM($\uparrow$)  & MACE(1e-4)($\downarrow$) & PSNR & SSIM  & MACE & PSNR & SSIM & MACE \\

\midrule
\multirow{2}{*}{TPDM\cite{lee2023improving}}  & 32.23 & 0.922 & \multirow{2}{*}{53.16} & 25.35 & 0.868 & \multirow{2}{*}{279.3} & 35.12 & 0.945 & \multirow{2}{*}{48.72} \\
&$\pm$ 0.0066 & $\pm$ 0.000300  &   &$\pm$ 0.2363 &$\pm$ 0.00112&   &$\pm$ 0.0154 &$\pm$ 0.00056 &  \\

\multirow{2}{*}{Ours-TPDM} & 33.24  & 0.944 & \multirow{2}{*}{44.63} & 25.26 & 0.882 & \multirow{2}{*}{268.3} & 36.24 & 0.961 & \multirow{2}{*}{38.49} \\
&$\pm$ 0.0298 & $\pm$ 0.000261 &  &$\pm$ 0.664 & $\pm$ 0.00357 &  &$\pm$ 0.0389 &$\pm$ 0.000249 & \\

\midrule
\multirow{2}{*}{TOSM\cite{li2024two}}  & 32.76 & 0.932 & \multirow{2}{*}{51.68} & 25.66 & 0.881 & \multirow{2}{*}{221.7} & 35.44 & 0.947 &  \multirow{2}{*}{46.32} \\
&$\pm$ 0.02157 & $\pm$ 0.000434 & &$\pm$ 0.2703 &$\pm$ 0.00132 &   &$\pm$ 0.0173 &$\pm$ 0.00073 & \\

\multirow{2}{*}{Ours-TOSM} & 33.30 & 0.945 & \multirow{2}{*}{42.89} & 25.24 & 0.882 & \multirow{2}{*}{200.8} & 36.51 & 0.963  & \multirow{2}{*}{36.67} \\
&$\pm$ 0.0296 & $\pm$ 0.000240 & &$\pm$ 0.668 & $\pm$ 0.00317& &$\pm$ 0.0364 &$\pm$ 0.000210 & \\

\midrule

\multirow{2}{*}{MADM\cite{averagingdiffusion}}  & 33.02 & 0.946 & \multirow{2}{*}{83.20} & 25.47 & 0.874  & \multirow{2}{*}{251.4} & 35.21 & 0.946  & \multirow{2}{*}{47.89} \\
&$\pm$ 0.0276 & $\pm$ 0.000436 & &$\pm$ 0.2573 &$\pm$ 0.00143 &   &$\pm$ 0.0165 & $\pm$ 0.00063 &  \\

\multirow{2}{*}{Ours-MADM} & 33.31 & 0.945 & \multirow{2}{*}{42.03} & 25.13 & 0.876 & \multirow{2}{*}{234.6} & 36.37 & 0.964 & \multirow{2}{*}{37.15} \\
&$\pm$ 0.0308 & $\pm$ 0.000278 &  &$\pm$ 0.667 &$\pm$ 0.00342&  &$\pm$ 0.0379 &$\pm$ 0.00226 &  \\
\bottomrule

\end{tabular}
}
\caption{Quantitative evaluation of Score-Fusion on BraTS dataset with uncertainty metrics. Our models' performance boost is significant, given low standard deviations. Our model can also estimate uncertainty better through the standard deviation obtained by inference multiple times.}

\label{tab:uncertain}
\vspace{-0.5cm}
\end{table*}
As demonstrated in Table~\ref{tab:uncertain}, all variants of our model exhibit lower MACE values compared to their respective baselines. This indicates that the standard deviation (std) predicted by our model, derived from multiple inferences, provides a more accurate estimation of the true error relative to the ground truth. Consequently, our model exhibits improved uncertainty awareness. For qualitative results in uncertainty awareness, we demonstrate our model's results with the uncertainty map and error map across various tasks and variants in Fig.~\ref{fig:SR1}~\ref{fig:SR2}~\ref{fig:MT1}~\ref{fig:MT2}~\ref{fig:cond2_1}~\ref{fig:cond2_2}~\ref{fig:TPDM_SR1}~\ref{fig:TPDM_SR2}. As shown in the figures, the uncertainty map aligns well with the actual error map, demonstrating decent uncertainty awareness for all models. Notably, our model usually has a higher uncertainty in modality translation tasks in Fig.~\ref{fig:MT1} and ~\ref{fig:MT2}. In the modality translation task, the $p(\bm{y} | \bm{x})$ should have a high variance in overall contrast. Our model outputs samples that are highly diverse in overall contrast, indicating that Score-Fusion is able to model the target 3D conditional distribution $p(\bm{y}  | \bm{x})$ better. In contrast, our baselines tend to output the mean estimation for overall contrast, demonstrating higher PSNR but limited capability of generating diverse and realistic results.

\section{Additional qualitative result}
\label{sec:visualization}
\begin{figure*}[h] \centering 
\includegraphics[width=\textwidth]{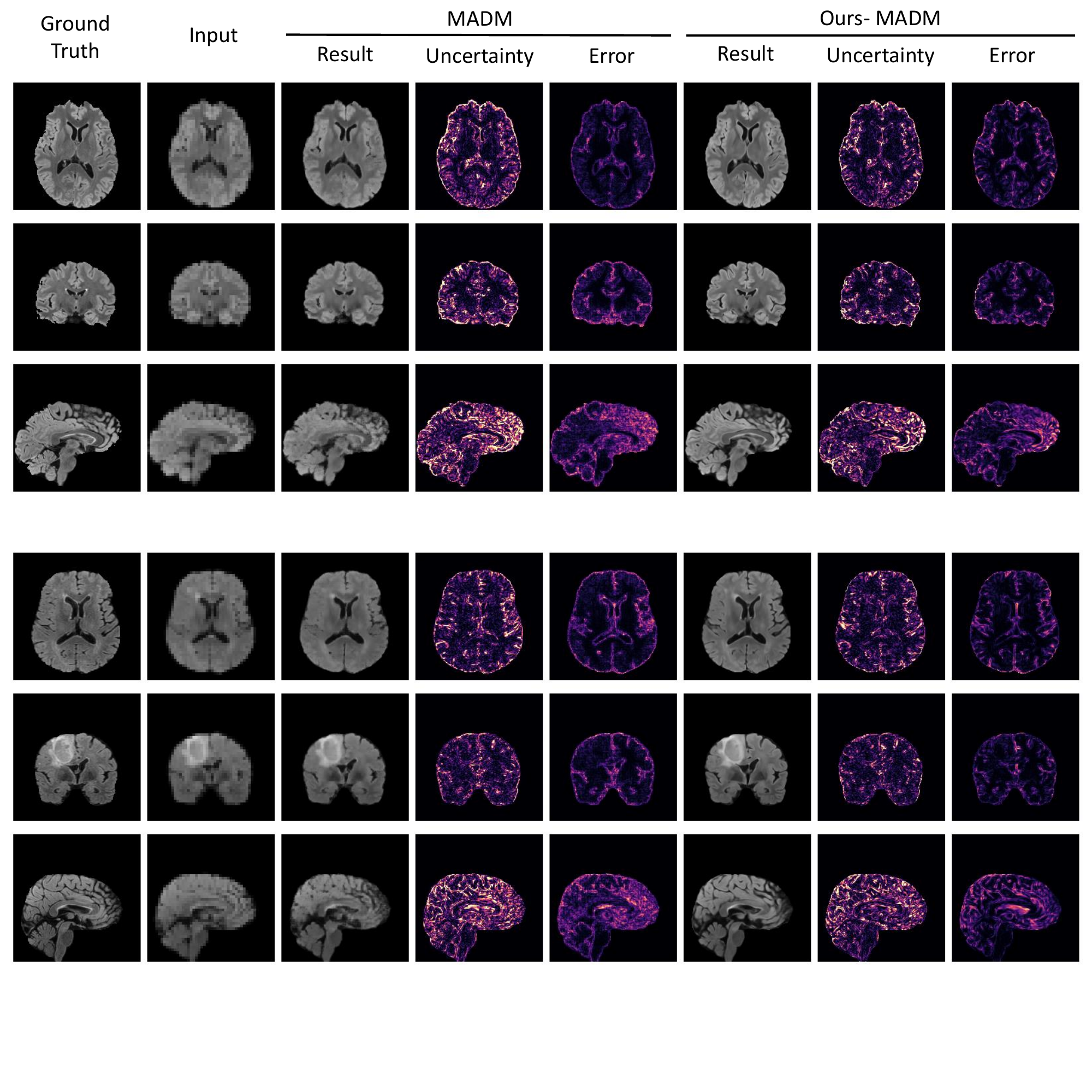} 
\vspace{-2cm}
\caption{Uncertainty awareness results on super-resolution task}
\label{fig:SR1}
\end{figure*}

\begin{figure*}[h] \centering 
\includegraphics[width=\textwidth]{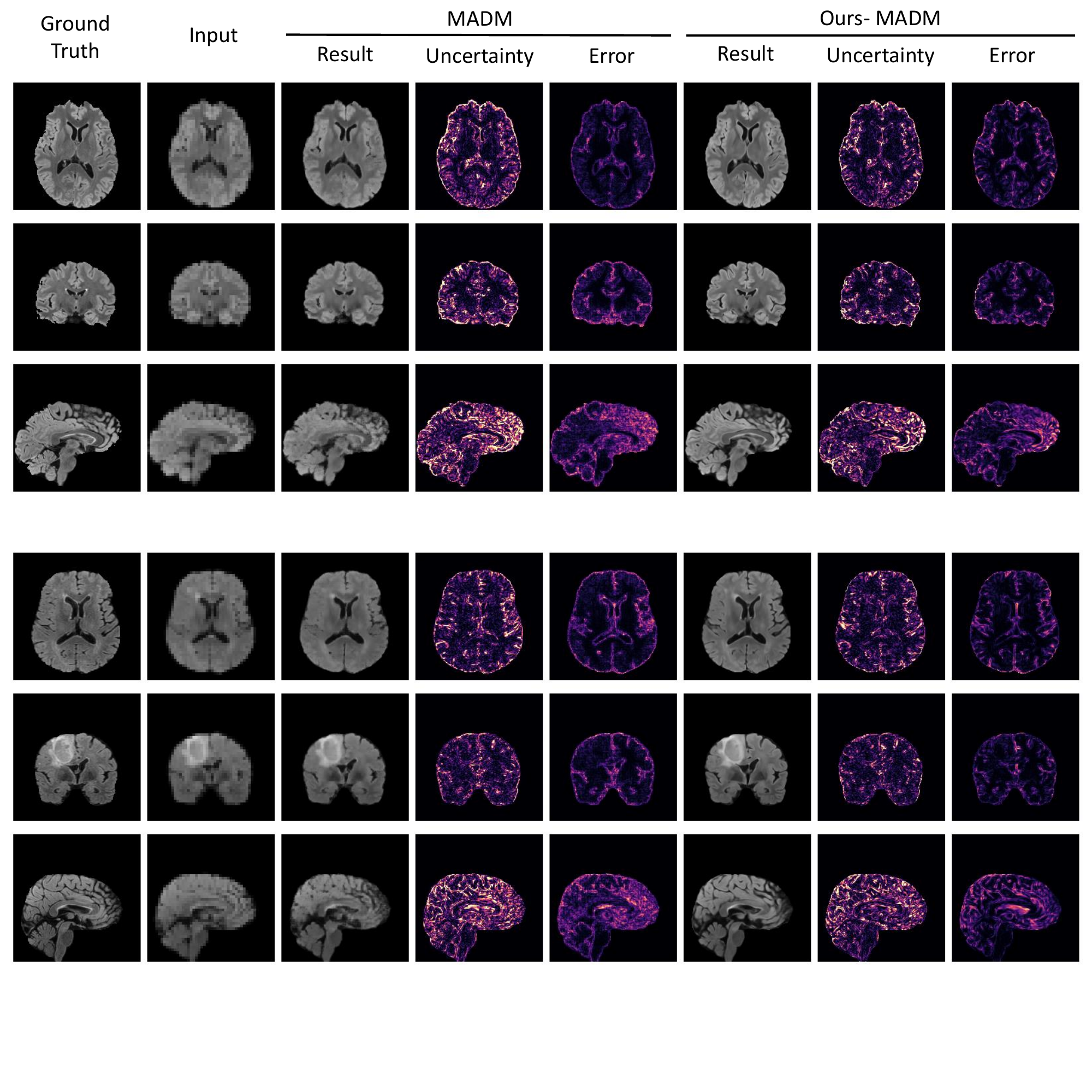} 
\vspace{-2cm}
\caption{Uncertainty awareness results on super-resolution task}
\label{fig:SR2}
\end{figure*}

\begin{figure*}[h] \centering 
\includegraphics[width=\textwidth]{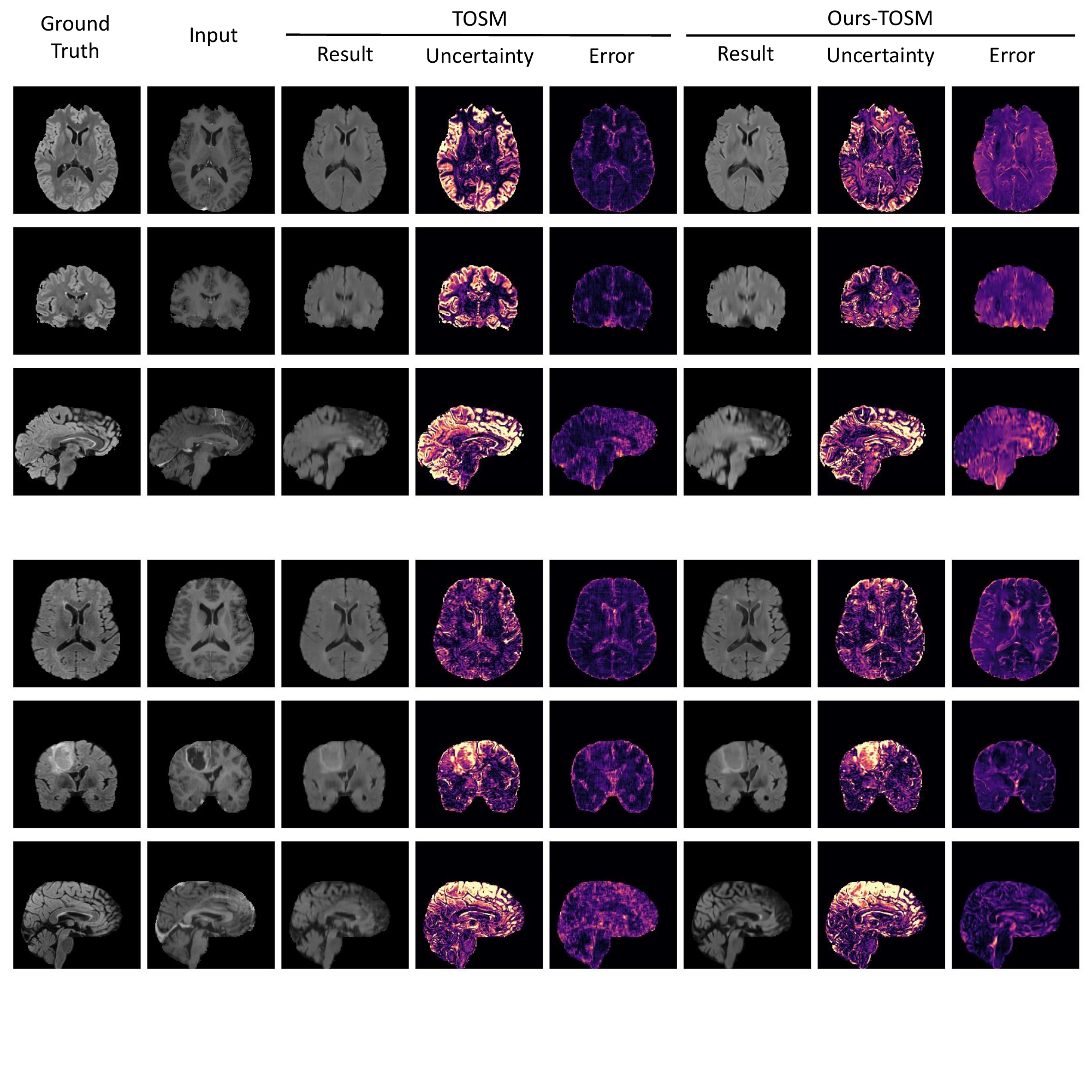} 
\vspace{-2cm}
\caption{Uncertainty awareness results on modality translation task}
\label{fig:MT1}
\end{figure*}

\begin{figure*}[h] \centering 
\includegraphics[width=\textwidth]{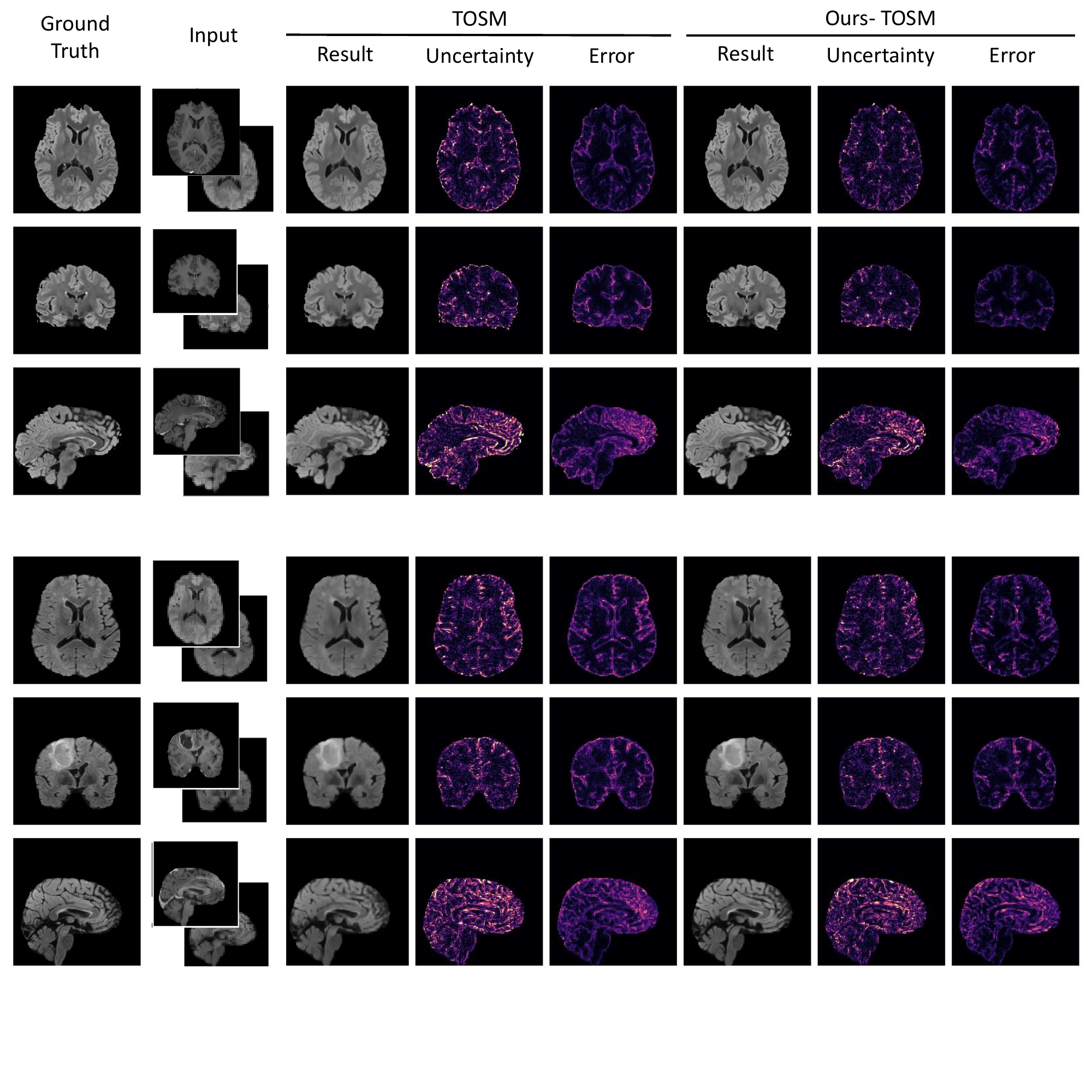} 
\vspace{-2cm}
\caption{Uncertainty awareness results given both conditions}
\label{fig:cond2_1}
\end{figure*}

\begin{figure*}[h] \centering 
\includegraphics[width=\textwidth]{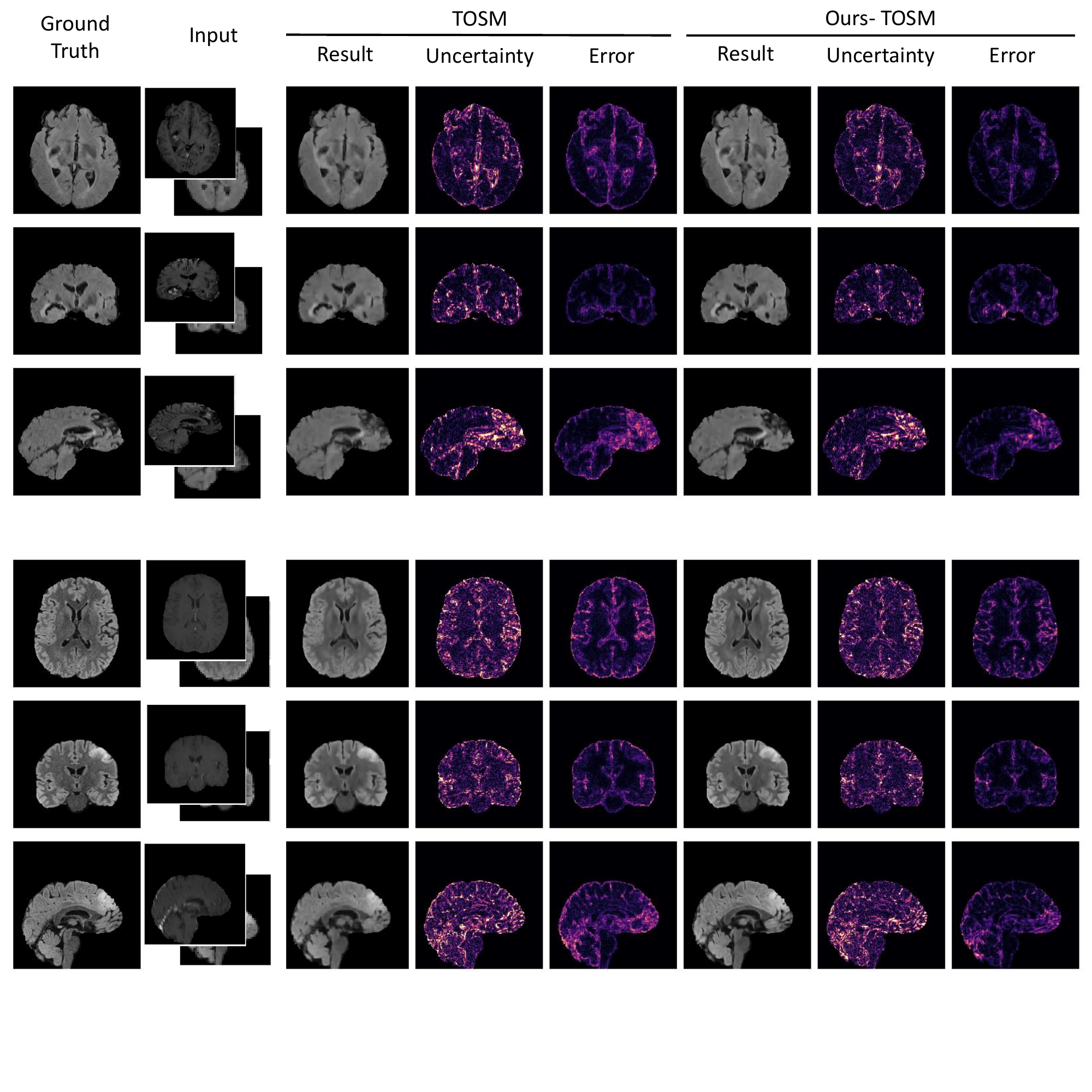} 
\vspace{-2cm}
\caption{Uncertainty awareness results given both conditions}
\label{fig:cond2_2}
\end{figure*}

\begin{figure*}[h] \centering 
\includegraphics[width=\textwidth]{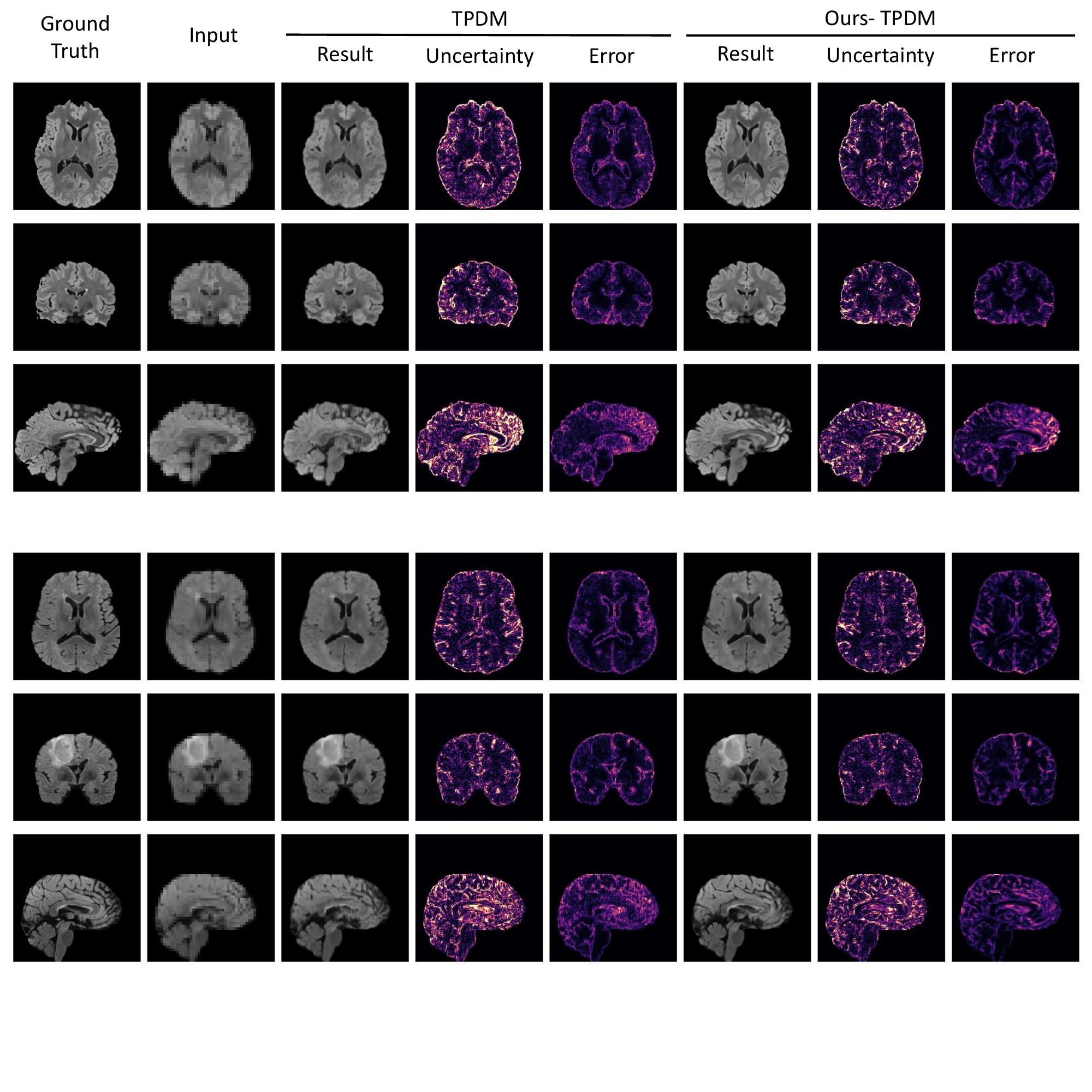} 
\vspace{-2cm}
\caption{Uncertainty awareness results on super-resolution for TPDM and Ours-TPDM}
\label{fig:TPDM_SR1}
\end{figure*}

\begin{figure*}[h] \centering 
\includegraphics[width=\textwidth]{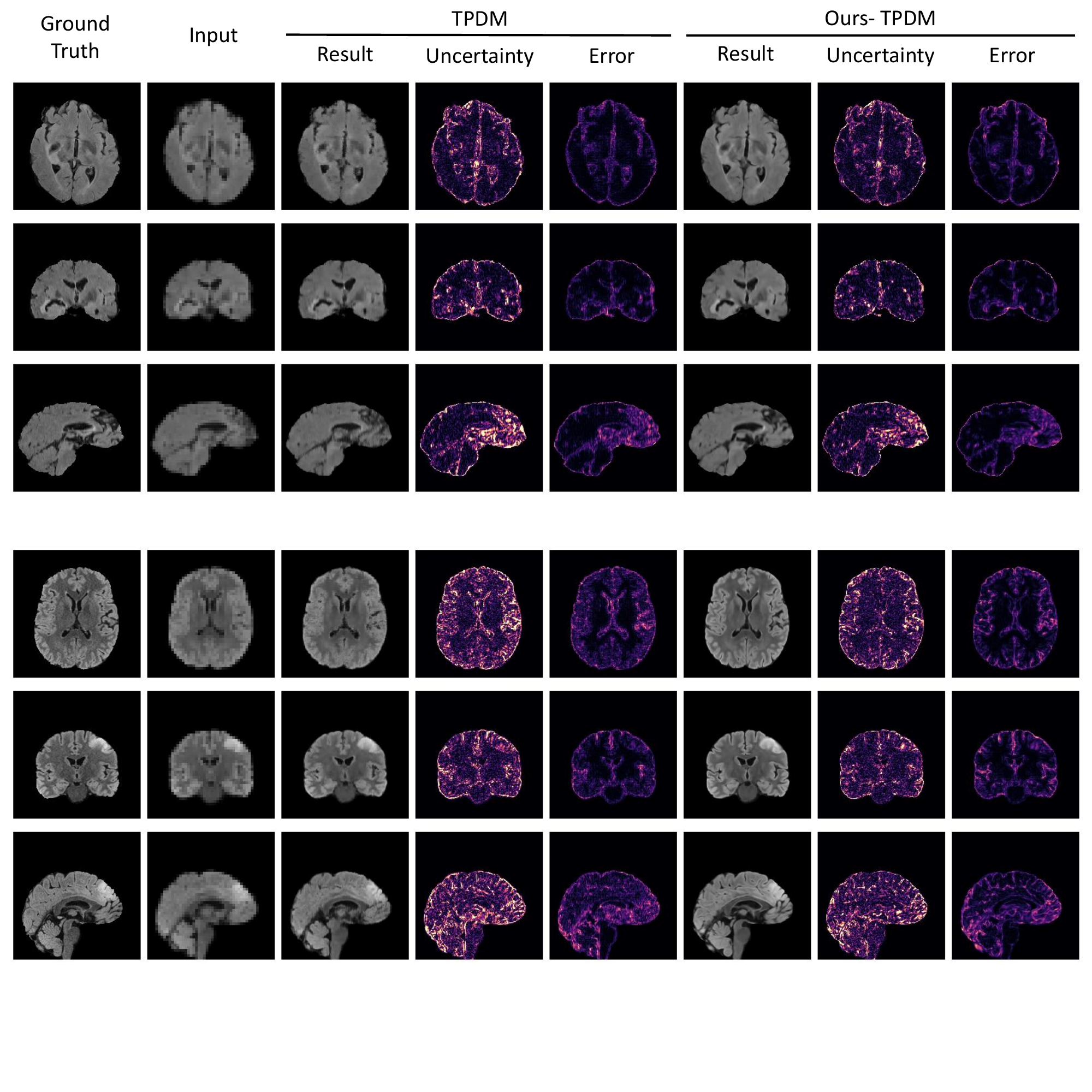} 
\vspace{-2cm}
\caption{Uncertainty awareness results on super-resolution for TPDM and Ours-TPDM}
\label{fig:TPDM_SR2}
\end{figure*}

We show results for the variants that show the best metrics. Namely, we show results for MADM and Ours-MADM in the super-resolution task in Fig.~\ref{fig:SR1} and ~\ref{fig:SR2}, and show results for TOSM and Ours-TOSM for the other two tasks in Fig.~\ref{fig:cond2_1}~\ref{fig:cond2_2}~\ref{fig:MT1} and ~\ref{fig:MT2}. We include more samples in all 3 views in Fig.~\ref{fig:TPDM_SR1} and Fig.~\ref{fig:TPDM_SR2} in super-resolution tasks in addition to Fig.~\ref{fig:zoomin}. Each figure contains 2 sample volumes, each of which contains visualizations in all three views in three rows. We show all results with uncertainty and error maps.

Similarly to Fig.~\ref{fig:zoomin}, we find that both MADM and TOSM demonstrate similar artifacts as TPDM in high-frequency details due to a direct averaging in the score function. In contrast, Score-Fusion consistently demonstrates better 3D consistency and realism across all views by introducing pixel-space 3D representation and networks to replace the weighted averaging in the score function space. For example, in the 6-th row of Fig.~\ref{fig:SR1}~\ref{fig:MT1}, and ~\ref{fig:cond2_1}, the results from baselines are blurry at the top left part of the brain, whereas Score-Fusion shows more smooth and consistent results.

\section{Result on HCP dataset}
\label{sec:HCP}
\begin{figure}[h] \centering 
\includegraphics[width=0.45\textwidth]{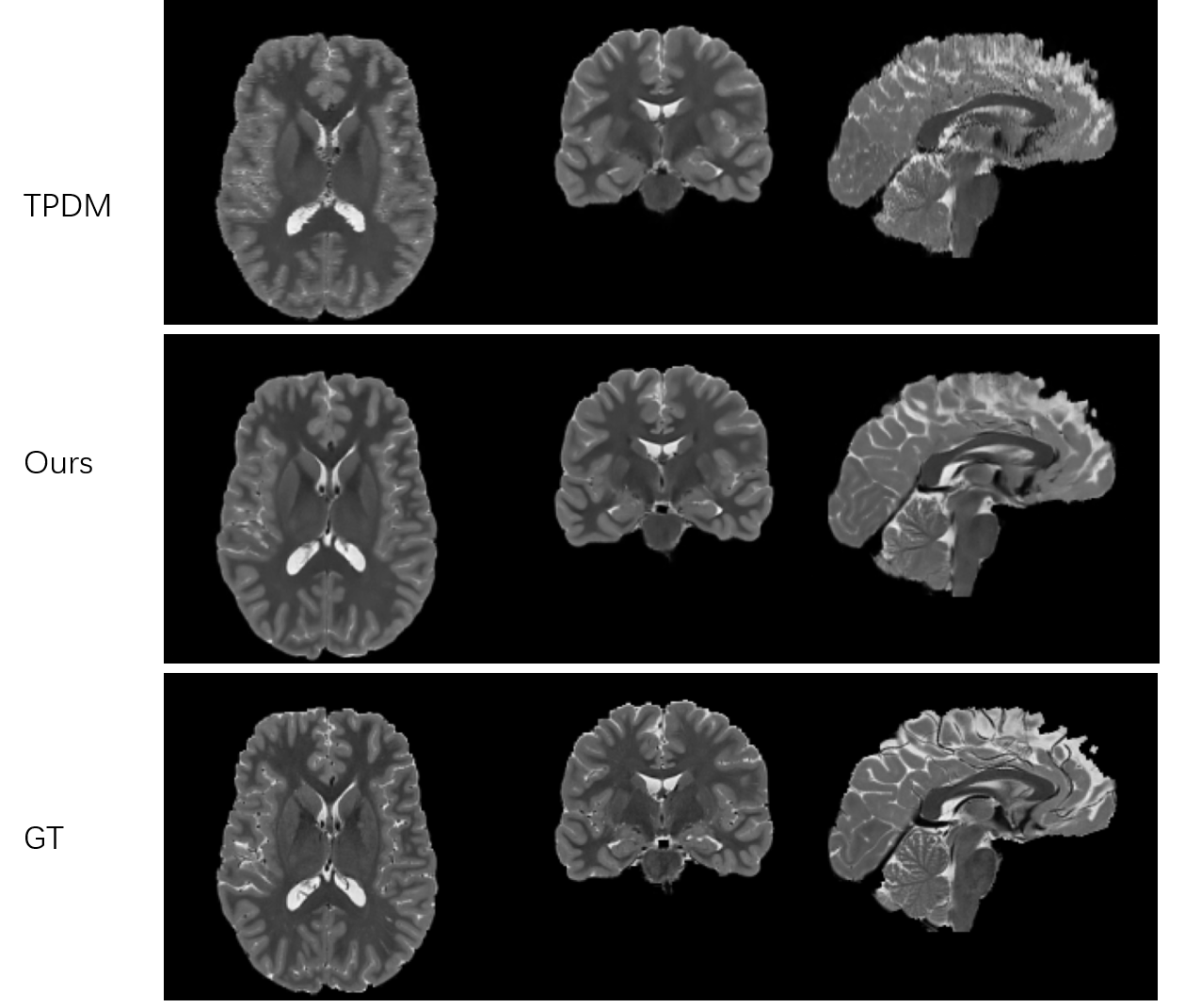} 
\caption{Qualitative results in HCP dataset. The input is a 4x4x4 downsampled version of the ground truth.}
\label{fig:HCP}
\end{figure}

We present our super-resolution results on the FLAIR modality in the HCP~\cite{HCP} dataset to show our model is generalizable across datasets. The HCP dataset consists of 1251 MRI volumes with a resolution of 192x152x152. In contrast to the BraTs dataset, HCP comprises healthy brains with no brain tumors. 
Experiments results in Tab.~\ref{tab:HCP} show that our model shows around 1.5 performance boost in PSNR. 
We also present the qualitative results in Fig.~\ref{fig:HCP}, including all 3 views. Again, Score-Fusion shows better 3D consistency and realism. For example, in the top-right part in the third view, our baseline demonstrates jittering and artifacts, while our model produces more realistic detail and smoother edges. 

\begin{table}[H]
\setlength{\tabcolsep}{4pt}
\small\centering
\scalebox{1.0}{
\begin{tabular}{ccccc}
\toprule    
Method & PSNR & SSIM & MMD & FID(1e-4)\\
\midrule
TPDM & 28.17 & 0.890 & 81.81 & 35.70 \\
Ours-TPDM & 29.62 & 0.914 & 67.96 & 22.12  \\

\bottomrule
\end{tabular}
}
\caption{Super-resolution result in HCP dataset. }
\label{tab:HCP}
\end{table}

\section{Ablation Studies}
\label{sec:ablation}
We provide an ablation study in Tab.~\ref{tab:abalation} on the key design elements for Score-Fusion, which includes: (1) \textbf{Feature merging}: The 2D models not only contribute their outputs but also pass their feature maps to the 3D model. (2) \textbf{Finetune}: We initially pre-train the model on 3D patches and then fine-tune it on the full volume to speed up training. (3) \textbf{Consistency}: Inspired by DPS~\cite{DPS} and score-SDE~\cite{scoreSDE}, we implement self-consistency projections at each denoising step. All these designs show performance gain in the super-resolution task. In addition, we benchmark the smaller variant in Sec.~\ref{sec:multi_res} for comparison, which shows a moderate performance drop compared to our best model.

\begin{table}[h]
\setlength{\tabcolsep}{4pt}
\small\centering
\scalebox{0.8}
{
\begin{tabular}{ccccccc}
\toprule    
Smaller model & Consistency & Finetune & Feature & PSNR & SSIM & MMD \\
\midrule
-- & -- & --  & -- & 32.83 & 0.935 & 25.45 \\
-- & \checkmark & --  & -- & 32.88 & 0.94 & 18.84 \\
-- & \checkmark & \checkmark & -- & 32.97 & 0.941 & 15.24 \\
-- & \checkmark & -- & \checkmark & 33.04 & 0.942 & 17.76 \\
-- & \checkmark & \checkmark & \checkmark & 33.24 & 0.944 & 13.77\\
 \checkmark & \checkmark & \checkmark & -- & 32.8 & 0.937 & 16.78\\
\bottomrule
\end{tabular}
}
\caption{Ablation studies of additional design elements in Score-Fusion.}
\vspace{-0.7cm}
\label{tab:abalation}
\end{table}

\section{Detailed model architecture}
\label{sec:architecture}
In this section, we show detailed model architecture for 2D, 3D, and the smaller variant of the 3D model in Tab.~\ref{tab:dif2D}, Tab.~\ref{tab:dif3D}, and Tab.~\ref{tab:small}, respectively. In addition, we show other related hyper-parameters in Tab.~\ref{tab:hyper}.
We modified the architecture of the 2D diffusion model from Palette~\cite{saharia2022palette} and the 3D models from med-ddpm~\cite{dorjsembe2024conditional}.

Given the differences in problem setting and dataset between our work and that of Palette, we conduct a comprehensive hyper-parameter search based on the super-resolution tasks. This search explores various configurations, including the number of channels, transformer layers, and learning rate, among others. The hyper-parameter search is conducted to optimize the performance of our baseline models, Palette2D, Palette3D, and Palette2.5D, in Tab.~\ref{tab:metrics}. While such a search could potentially enhance the performance of our proposed model, we do not perform a hyper-parameter search to optimize the performance of Score-Fusion, TPDM, TOSM, and MADM. This practice ensures a fair comparison between our model and their corresponding baselines, TPDM, TOSM, and MADM. Moreover, this shows that our model can be a plug-in-and-play mechanism for existing pre-trained 2D and 3D model architecture.

\begin{table}[H]
\centering
\caption{Other hyper-parameters}
\label{tab:hyper}
\begin{tabular}{ccc}
\toprule
\textbf{Parameter} & \textbf{2D Network} & \textbf{3D Network} \\ \midrule
Batch size & 4 & 1 \\ 
Diffusion steps & 1000 & 1000 \\ 
Inference steps (DDIM) & 50 & 50 \\ 
Noise scheduler & Linear & Linear \\ 
Learning rate & 0.00005 & 0.0001 \\ 
Optimizer & Adam & Adam \\ \bottomrule
\end{tabular}
\end{table}

\section{Training and inference speed}
\label{sec:speed}
We present training inference speed in Tab.~\ref{tab:speed}. All experiments are done with RTX A100-40GB GPU.
Since Score-Fusion needs to train an additional model on top of the baselines, our training time is inevitably higher. We need 16 GPU days to train our 3D models, which results in a 16-day increase in training time for most model variants compared to their corresponding baselines. Our models are also relatively slower in inference since we need to perform inference for an additional 3D model. However, as mentioned in Sec.~\ref{sec:Introduction}, the 3D model is naturally limited in size due to computational challenges in training. Therefore, 3D inference is more efficient than slice-wise 2D inference. As a result, the increase in inference time is significantly smaller than in training. As shown in Tab.~\ref{tab:speed}, our 3D model is around 30\% faster than one 2D model and, therefore, leads to a 36\% increase in inference time for Ours-TPDM and 26\% for Ours-MADM and Ours-MADM. 

Moreover, we find that the TPDM-based models are significantly faster than other variants of the models. Given the advantage of computational efficiency, we use TPDM and Ours-TPDM as our main variables for the model and the baseline. Furthermore, to perform a more complete ablation study, the smaller 3D model decreases the inference and training time of the 3D model by 75\% while showing a consistent performance boost over TPDM and a moderate performance drop compared to Ours-TPDM as shown in Tab.~\ref{tab:metrics}. Moreover, compared to 3D Palette baseline, our model effectively decreased 3D training from 120 GPU days to 16/4 days, addressing the computational challenge of 3D diffusion training.

\begin{table}[h]
\centering
\caption{Training and Inference time for each model, GPUs are A100 with 40G memory.}
\label{tab:speed}
\begin{tabular}{ccc}
\toprule
\multirow{2}{*}{\textbf{Time}} & \textbf{Training time } & \textbf{Inference time } \\ 
& (GPU days) & (minutes per volume) \\
\midrule

2D Palette & 8 & 0.85\\ 
2D I2SB & 5 & 4.58\\ 
3D Pix2pix  & 6 & 0.0398\\ 
3D Unet  & 6 & 0.0398\\ 
3D Palette & 120 & 0.6 \\ 
TPDM & 16 & 1.72 \\ 
Ours-TPDM & 32 & 2.34 \\ 
Ours-TPDM-small & 20 & 1.92 \\
TOSM & 24 & 2.55 \\ 
Ours-TOSM & 40 & 3.23 \\ 
MADM & 36 & 2.76\\ 
Ours-MADM & 52 & 3.56 \\ 

\bottomrule
\end{tabular}

\end{table}

\section{Details for consistency projection}
\label{sec:consistency}
In this section, we provide the exact definition and detail for self-consistency projection mentioned in Sec.~\ref{sec:pre}. In this work, we address the inverse problem using a diffusion model with consistency projections. The goal is to recover a high-resolution image, $\bm{y}$, from its low-resolution observation $\bm{x}$, which is obtained through a linear degradation process. Specifically, the degradation process is modeled as: $\bm{x} = A \bm{y}$.

In the 3D case, the degradation operator $A$ represents a downsampling operation that reduces the resolution of a volume $\bm{y}$ by a factor of 4 along each spatial dimension (x, y and z) and resizes it back to the original resolution. This means that each voxel in the low-resolution volume $\bm{x}$ corresponds to the average of a 
[4×4×4] region in the high-resolution volume $\bm{y}$.
Specifically, let $\bm{y}, \bm{x} \in \mathbb{R}^{b_1 \times b_2 \times b_3}$. The operator matrix $A \in \mathbb{R}^{b_1 \times b_2 \times b_3,b_1 \times b_2 \times b_3}$ downscales the high-resolution volume $\bm{y}$ into the low-resolution volume $\bm{x}$ by averaging over [4x4x4] blocks of voxel of $\bm{y}$. Therefore, $A$ is a sparse matrix where each non-zero entry corresponds to the average of a block of [4x4x4] voxels in $\bm{y}$ being averaged to form a block of voxel in $\bm{x}$. Therefore, $A$ is $\frac{1}{64}$ for the places where $\bm{x} $ and $ \bm{y}$ belong to the same block, and $A$ would be 0 elsewhere:
\begin{equation}
\begin{aligned}
    & A[(i,j,k), (p,q,r)] = \frac{1}{64} & \text{ if } \bm{x}(i,j,k),\bm{y}(p,q,r) \in block \\
    & A[(i,j,k), (p,q,r)] = 0 & \text{ Otherwise }
\end{aligned}
\end{equation}
Since we are doing average over [4x4x4], $\bm{x}(i,j,k)$ and $\bm{y}(p,q,r)$ are in the same block if and only if $ i // 4 == p // 4 $, $ j // 4 == q // 4 $, and $ k // 4 == r // 4 $. 

In our diffusion process, we use $\bm{\hat{y}}_0(t) \gets \bm{\hat{y}}_0(t) - A^T{(AA^T)}^{-1}(A\bm{\hat{y}}_0(t) - x)$ to make every of our mean prediction of $\hat{y}_0$ a plausible estimation with $\bm{x} = A\hat{y}_0(t)$

To compute matrix multiplication more efficiently in a super-resolution setting, we actually use $\hat{y}_0(t) \gets \hat{y}_0(t) - (A\hat{y}_0(t) - x)$ in our code. This works in the average pooling downsample because $AA\bm{y} = A\bm{y}$ since $A$ represents the degradation process composed of average pooling followed by resizing the image back to its original resolution.

\section{Details in downstream task}
\label{sec:downstream_detail}   

In Section~\ref{sec:downstream}, we evaluate tumor segmentation performance using three types of FLAIR inputs: the ground truth FLAIR modality, 4x downsampled FLAIR modality (as described in Section~\ref{exp:experimentalsetup}), and the 4x super-resolution FLAIR prediction on the downsampled FLAIR. Accurate tumor segmentation is crucial in medical imaging, and its performance heavily relies on the quality of the input data. It requires High-quality inputs for precise localization and delineation of tumor boundaries, while, depending on its degradation level, the degraded inputs could significantly lower segmentation accuracy and reliability. We use a robust pre-trained segmentation model, SwinUNet~\cite{SWINUnet}, which takes four modalities (T1, T1ce, T2, and FLAIR) as input. For this downstream task, our objective is to assess how well the models can recover segmentation performance when working with degraded inputs. Segmentation is performed with other modalities with ground truth inputs and a FLAIR input from the ground truth FLAIR, downsampled FLAIR, or the model-predicted FLAIR. Note that because there is no degraded FLAIR modality available in the modality translation task, only dice scores are reported. For other tasks, including the super-resolution and both condition tasks, performance is measured using two metrics: (1) \textbf{Dice Score}, the primary metric of the segmentation model, and (2) \textbf{Recovery Rate}, a measure of how well model predictions improve upon degraded FLAIR inputs. The recovery rate is calculated as: 
\[
\text{Recovery Rate} = \frac{\text{Prediction} - \text{Downsample}}{\text{Ground Truth} - \text{Downsample}}
\]
where \textbf{Prediction} refers to the segmentation performance using predicted FLAIR, \textbf{Downsample} is the performance with downsampled FLAIR, and \textbf{Ground Truth} is the performance with ground truth FLAIR.

\begin{figure*}[h]
    \centering
    \includegraphics[width=1\columnwidth,height=1\textheight,keepaspectratio]{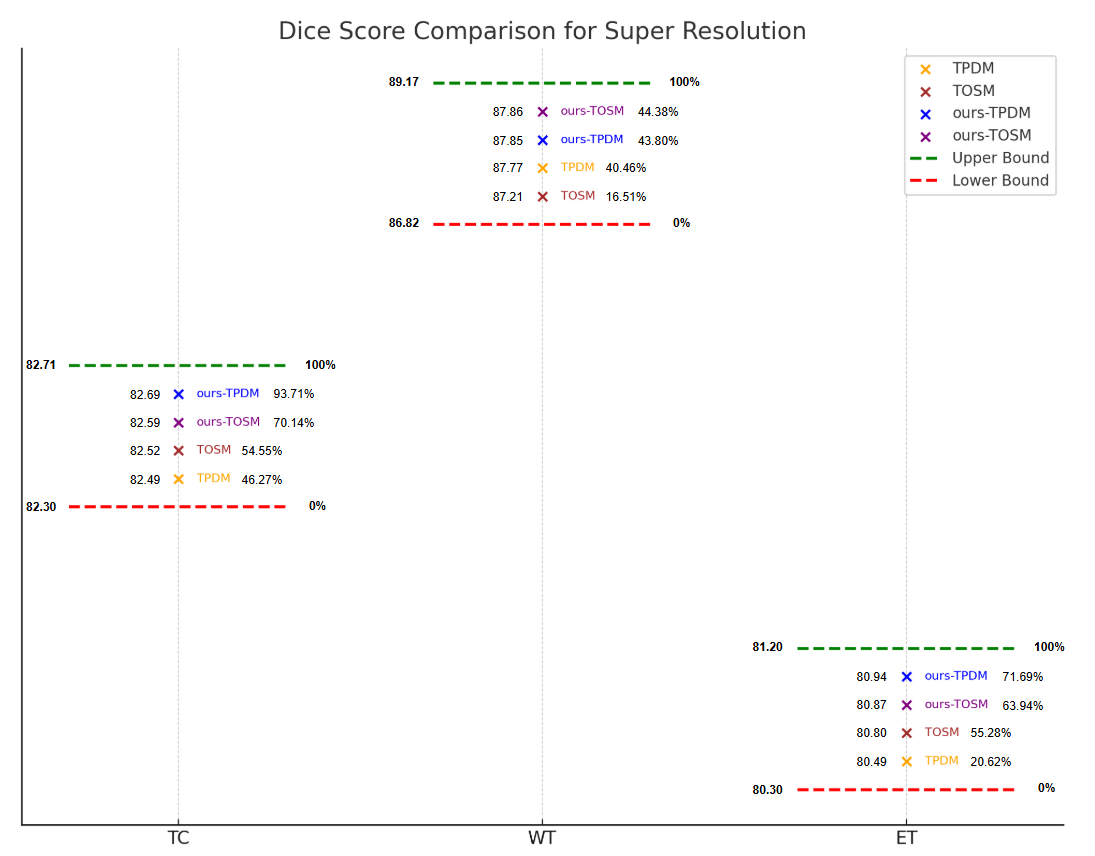}
\caption{Comparison of Dice scores and recovery rates for super-resolution. The value on the left represents the Dice score, while the value on the right represents the recovery rate.}
\label{fig:downstream_dice_sr}
\end{figure*}

\begin{figure*}[h]
    \centering
    \includegraphics[width=1\columnwidth,height=1\textheight,keepaspectratio]{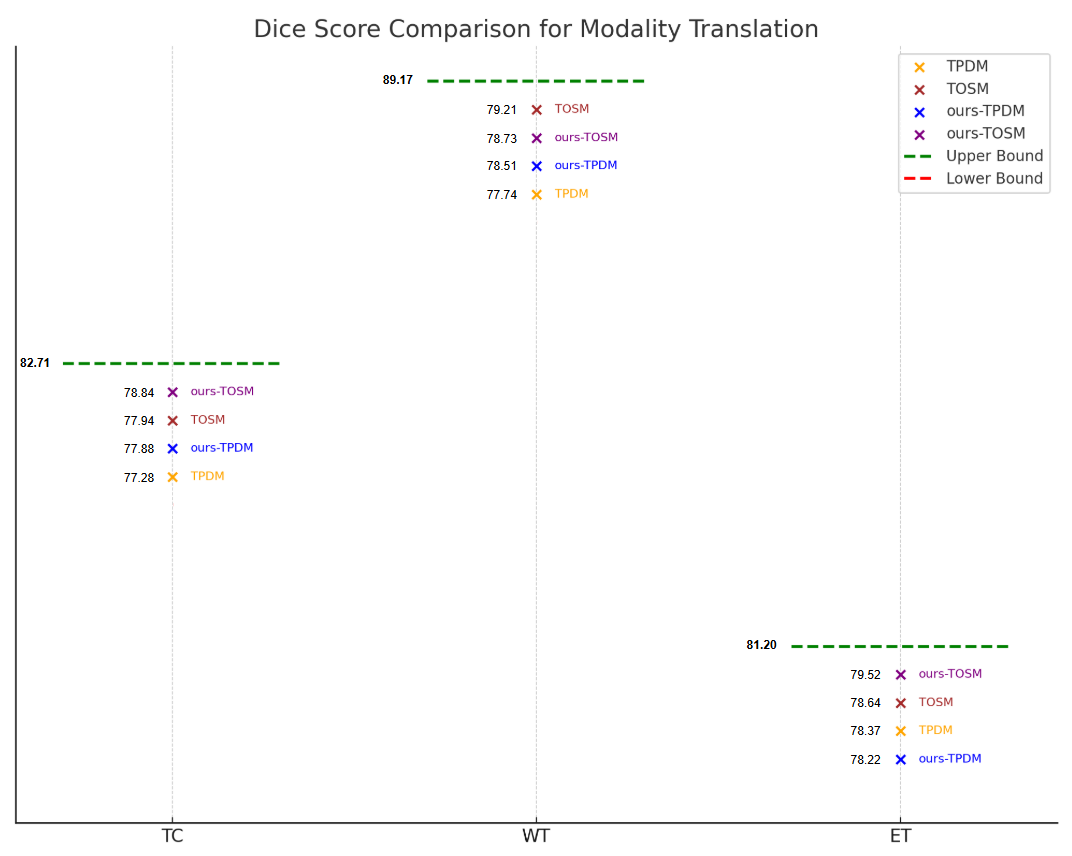}
\caption{Comparison of Dice scores and recovery rates for modality translation. The value on the left represents the Dice score.}
\label{fig:downstream_dice_mt}
\end{figure*}

\begin{figure*}[h]
    \centering
    \includegraphics[width=1\columnwidth,height=1\textheight,keepaspectratio]{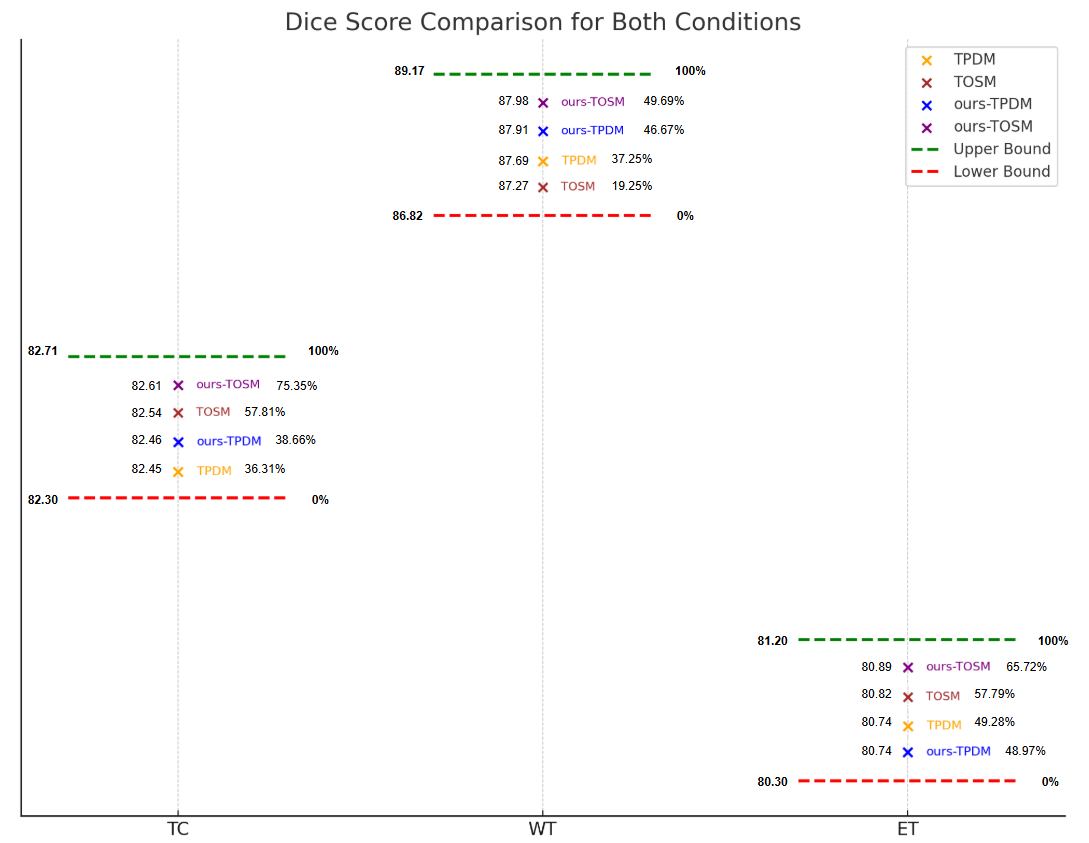}
\caption{Comparison of Dice scores and recovery rates for both conditions. The value on the left represents the Dice score, while the value on the right represents the recovery rate.}
\label{fig:downstream_dice_both}
\end{figure*}

Fig.~\ref{fig:downstream_dice_sr},~\ref{fig:downstream_dice_mt},~\ref{fig:downstream_dice_both} illustrate the Dice score and Recovery rate comparisons across tumor categories. Dashed lines represent the lower and upper bounds. They show that segmentation performance with the predicted FLAIR modality from Score-Fusion-based models outperforms other methods, as Score-Fusion-based models are constantly positioned higher than others.

\begin{figure*}[h] \centering 
\includegraphics[width=\textwidth]{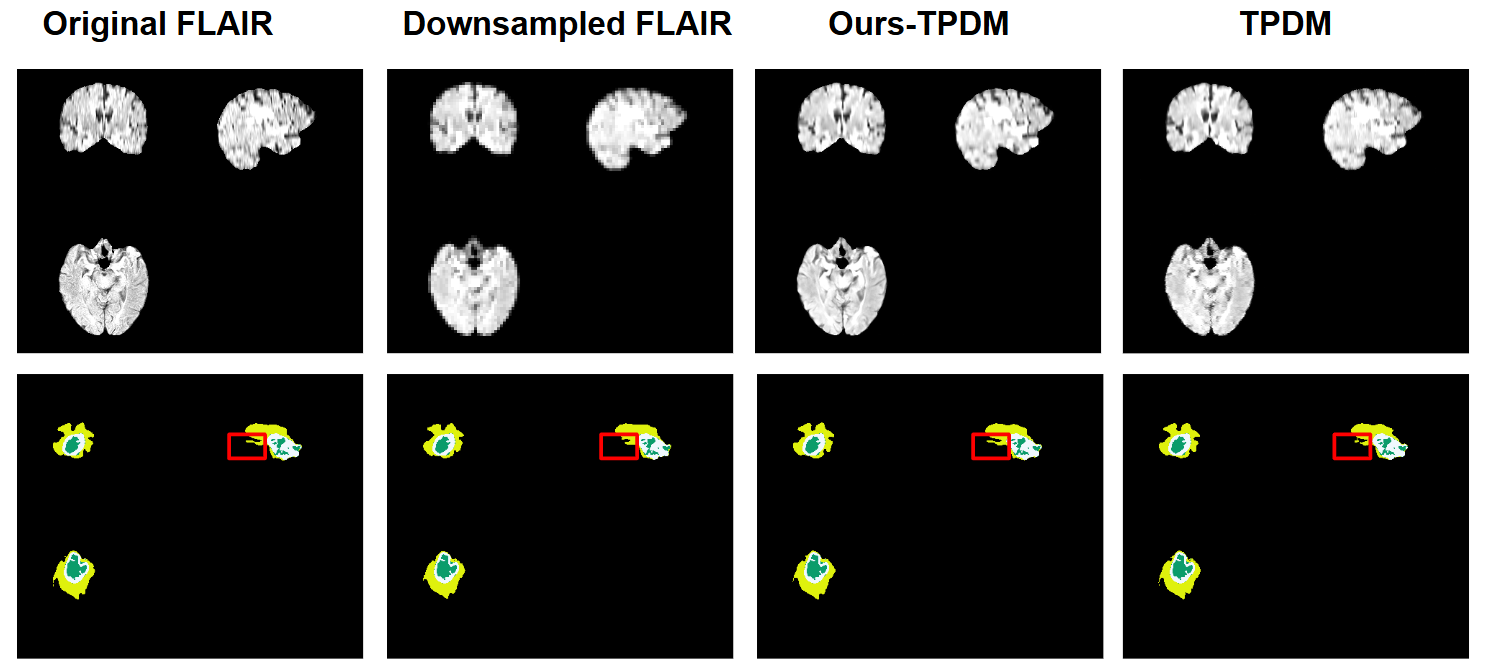} 
\caption{Qualitative results for the downstream task, tumor segmentation, in super-resolution task. }
\label{fig:downstream_SR1}
\end{figure*}

\begin{figure*}[h] \centering 
\includegraphics[width=\textwidth]{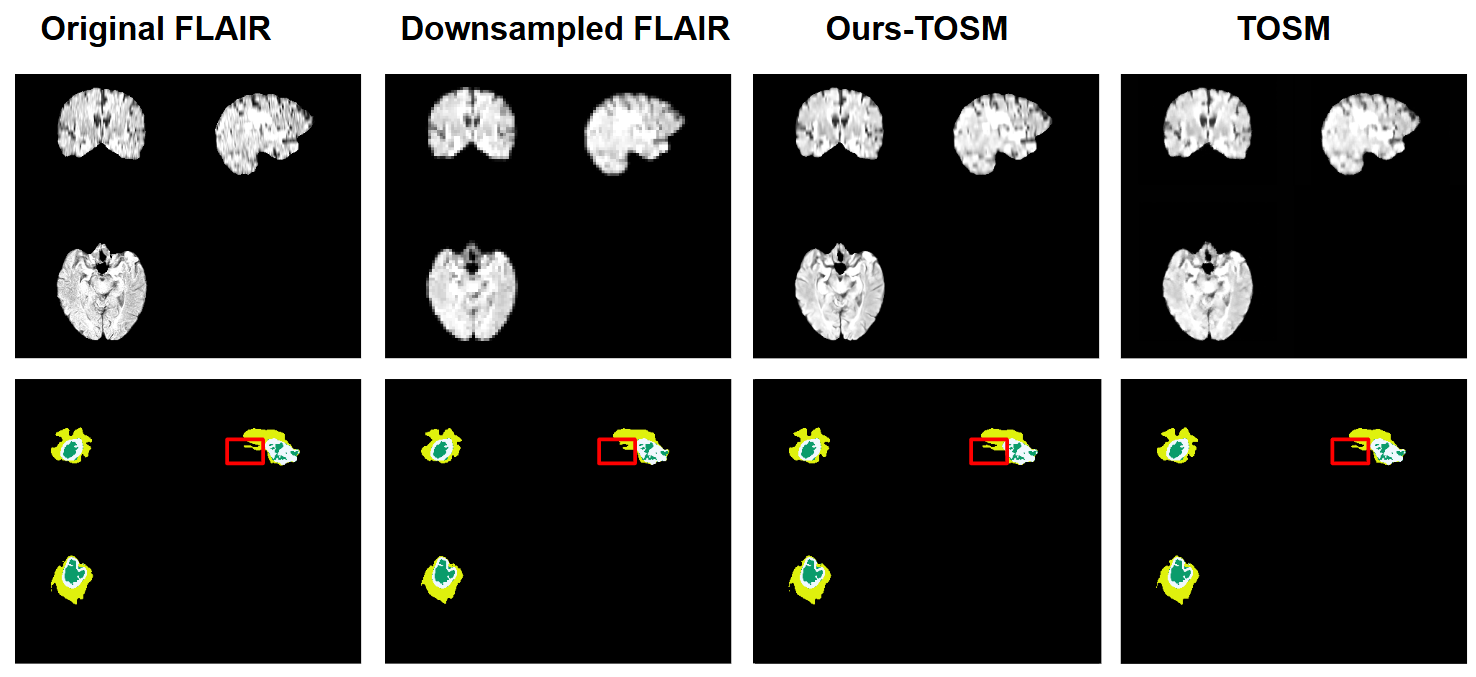} 
\caption{Qualitative results for the downstream task, tumor segmentation, in super-resolution task}
\label{fig:downstream_SR2}
\end{figure*}

\begin{figure*}[h] \centering 
\includegraphics[width=0.8\textwidth]{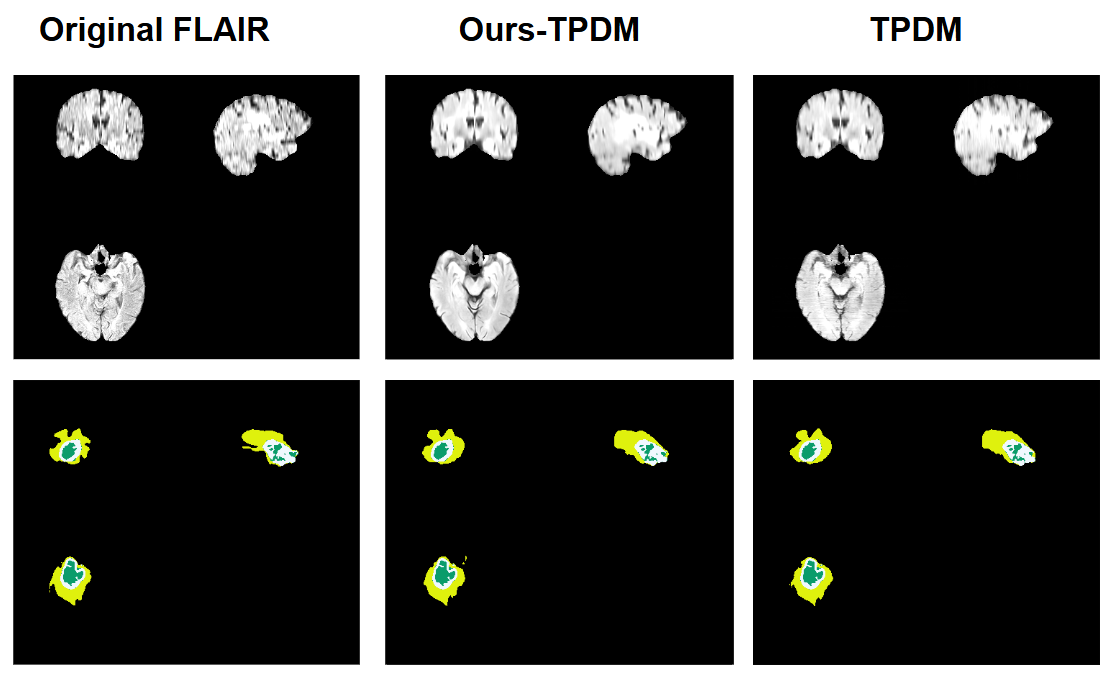} 
\caption{Qualitative results for the downstream task, tumor segmentation, in modality translation task}
\label{fig:downstream_MT1}
\end{figure*}

\begin{figure*}[h] \centering 
\includegraphics[width=0.8\textwidth]{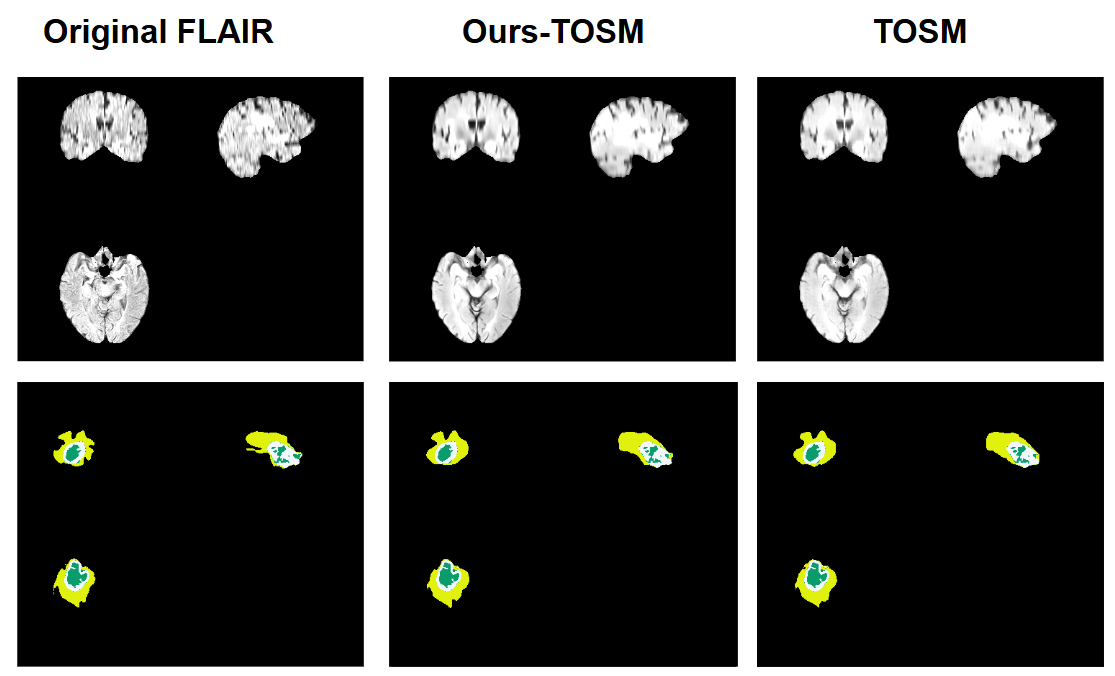} 
\caption{Qualitative results for the downstream task, tumor segmentation, in modality translation task}
\label{fig:downstream_MT2}
\end{figure*}

\begin{figure*}[h] \centering 
\includegraphics[width=\textwidth]{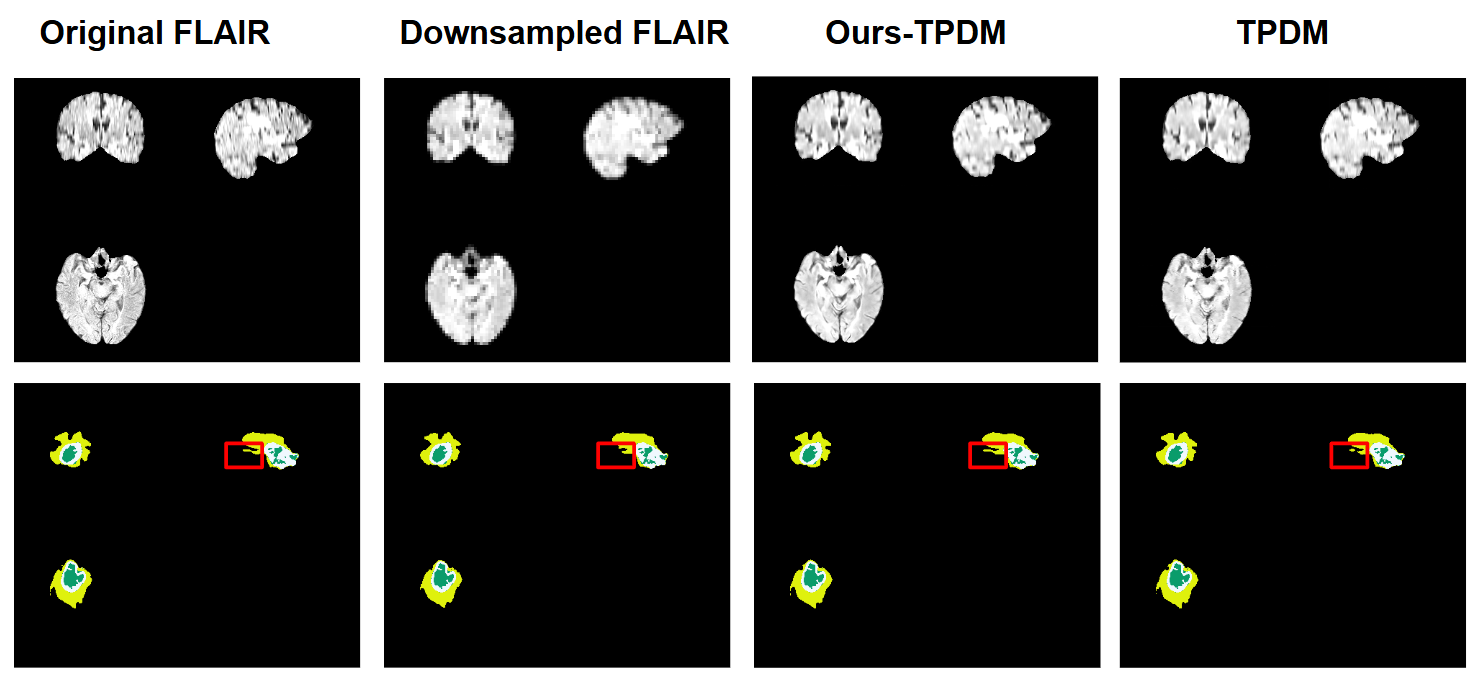} 
\caption{Qualitative results for the downstream task, tumor segmentation, in both-condition task}
\label{fig:downstream_2cond1}
\end{figure*}

\begin{figure*}[h] \centering 
\includegraphics[width=\textwidth]{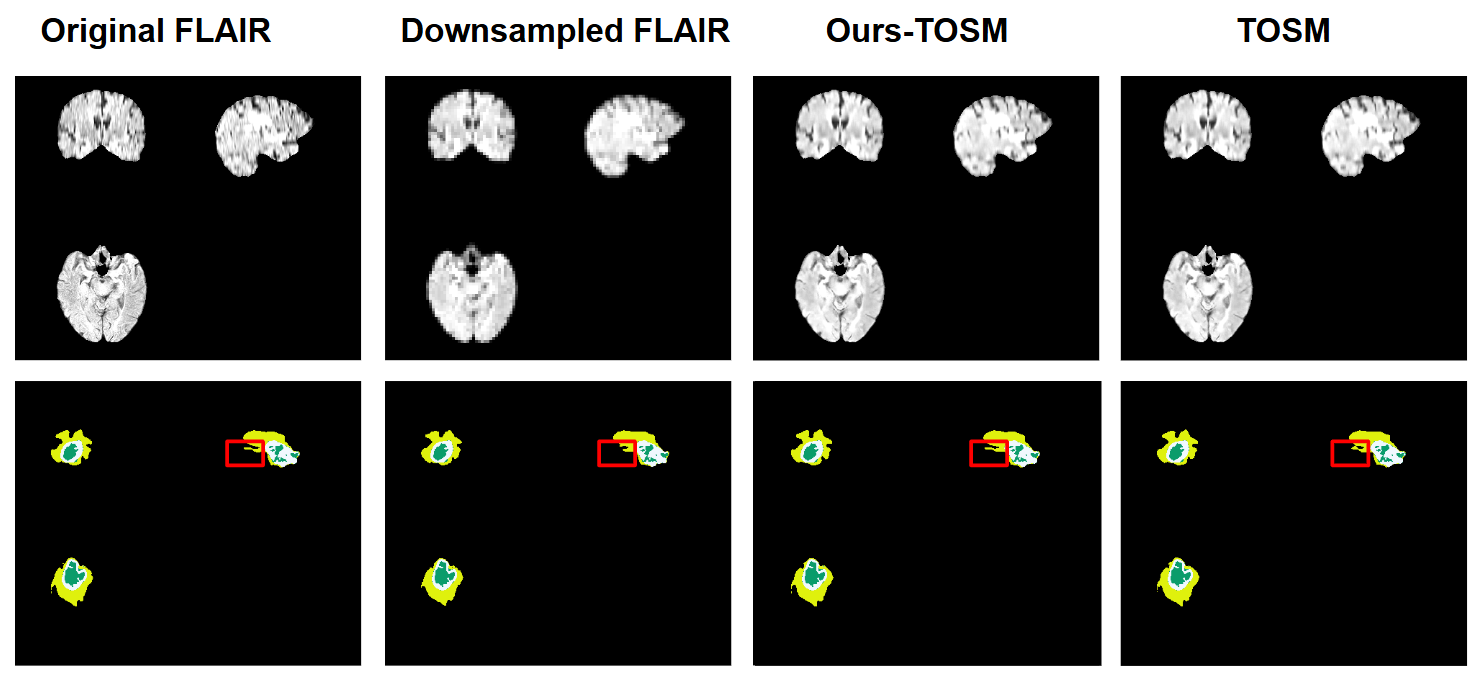} 
\caption{Qualitative results for the downstream task, tumor segmentation, in both-condition task}
\label{fig:downstream_2cond2}
\end{figure*}

We also show qualitative results in the tumor segmentation task, on TPDM, TOSM, and Score-Fusion built based on these two models. Fig~\ref{fig:downstream_SR1} and Fig~\ref{fig:downstream_SR2} show the results in super-resolution, Fig~\ref{fig:downstream_MT1} and Fig~\ref{fig:downstream_MT2} show the results in modality translation, and Fig~\ref{fig:downstream_2cond1} and Fig~\ref{fig:downstream_2cond2} show the results given both conditions. In Fig~\ref{fig:downstream_SR1},~\ref{fig:downstream_2cond2},~\ref{fig:downstream_2cond1}, and~\ref{fig:downstream_2cond2}, in the sagittal plane, we can observe that our models help segmentation model capture a branch coming out of the whole tumor, indicated by a red bounding box. This branch is only partially captured or entirely missed in predictions from other models. This shows that our model yields more precise predictions, allowing the tumor segmentation model to delineate the entire tumor boundary more accurately.

\begin{table*}[!t]\centering
\caption{Architecture for 2D diffusion model. Each ResnetBlock consists of 3 conv2D layers of the same channel and a skip connection. All ResnetBlocks are used with time embedding.}
\begin{tabular}{|c|c|c|c|c|c|c|}
\hline
\multicolumn{2}{|c|}{layers} & \multicolumn{5}{|c|}{parameters} \\
\hline
input & Conv3d & \multicolumn{5}{|c|}{in\_ch: 5, out\_ch: 64, kernel: 3x3, stride: 1, pad: 1 } \\
\hline
\multirow{3}*{Time\_Embed} &  Linear & \multicolumn{5}{|c|}{in\_ch:64, out\_ch: 256 } \\
~ &  Activateion & \multicolumn{5}{|c|}{Swish} \\
~ &  Linear & \multicolumn{5}{|c|}{in\_ch:256, out\_ch: 256 } \\
\hline

\multirow{4}*{Downsample\_block\_1} &  ResnetBlock & \multicolumn{5}{|c|}{in\_ch:64, out\_ch: 64 } \\
~ &  ResnetBlock & \multicolumn{5}{|c|}{in\_ch:64, out\_ch: 64 } \\
~ &  Downsample(Conv3d) & \multicolumn{5}{|c|}{in\_ch:64, out\_ch: 64,kernel:3x3, stride:2) } \\
\hline
\multirow{3}*{Downsample\_block\_2} &  ResnetBlock & \multicolumn{5}{|c|}{in\_ch:64, out\_ch: 128 } \\
~ &  ResnetBlock & \multicolumn{5}{|c|}{in\_ch:128, out\_ch: 128 } \\
~ &  Downsample(Conv3d) & \multicolumn{5}{|c|}{in\_ch:128, out\_ch: 128,kernel:3x3, stride:2 } \\
\hline
\multirow{3}*{Downsample\_block\_3} &  ResnetBlock & \multicolumn{5}{|c|}{in\_ch:128, out\_ch: 256 } \\
~ &  ResnetBlock & \multicolumn{5}{|c|}{in\_ch:256, out\_ch: 256 } \\
~ &  Downsample(Conv3d) & \multicolumn{5}{|c|}{in\_ch:256, out\_ch: 256,kernel:3x3, stride:2 } \\
\hline
\multirow{2}*{Downsample\_block\_4} &  ResnetBlock & \multicolumn{5}{|c|}{in\_ch:256, out\_ch: 512 } \\
~ &  ResnetBlock & \multicolumn{5}{|c|}{in\_ch:512, out\_ch: 512 } \\
\hline
\multirow{3}*{Middle} &  ResnetBlock & \multicolumn{5}{|c|}{in\_ch:512, out\_ch: 512 } \\
~ &  ResnetBlock & \multicolumn{5}{|c|}{in\_ch:512, out\_ch: 512 } \\
\hline
\multirow{3}*{Upsample\_block\_1} &  ResnetBlock & \multicolumn{5}{|c|}{in\_ch:512, out\_ch: 512 } \\
~ &  ResnetBlock & \multicolumn{5}{|c|}{in\_ch:512, out\_ch: 512 } \\
~ &  Upsample & \multicolumn{5}{|c|}{Conv3d and F.interpolate} \\
\hline
\multirow{3}*{Upsample\_block\_2} &  ResnetBlock & \multicolumn{5}{|c|}{in\_ch:512, out\_ch: 256 } \\
~ &  ResnetBlock & \multicolumn{5}{|c|}{in\_ch:256, out\_ch: 256 } \\
~ &  Upsample & \multicolumn{5}{|c|}{Conv3d and F.interpolate} \\
\hline
\multirow{3}*{Upsample\_block\_3} &  ResnetBlock & \multicolumn{5}{|c|}{in\_ch:256, out\_ch: 128 } \\
~ &  ResnetBlock & \multicolumn{5}{|c|}{in\_ch:128, out\_ch: 128 } \\
~ &  Upsample & \multicolumn{5}{|c|}{Conv3d and F.interpolate} \\
\hline
\multirow{2}*{Upsample\_block\_4} &  ResnetBlock & \multicolumn{5}{|c|}{in\_ch:128, out\_ch: 64 } \\
~ &  ResnetBlock & \multicolumn{5}{|c|}{in\_ch:64, out\_ch: 64 } \\
\hline

\multirow{3}*{Out} &  Normalize & \multicolumn{5}{|c|}{64} \\

~ &  Activation & \multicolumn{5}{|c|}{nn.SiLU} \\

~ &  Conv3d & \multicolumn{5}{|c|}{in\_ch:64, out\_ch: 1, kernel: 3x3, stride: 1, pad: 1} \\
\hline

\end{tabular}
\label{tab:dif2D}
\end{table*}

\begin{table*}[!t]\centering
\caption{Architecture for 3D diffusion model. Each ResnetBlock consists of 2 conv3D layers of the same channel and a skip connection. All ResnetBlocks are used with time embed with an embedding layer, as well as gradient checkpoint}
\begin{tabular}{|c|c|c|c|c|c|c|}
\hline
\multicolumn{2}{|c|}{layers} & \multicolumn{5}{|c|}{parameters} \\
\hline
input & Conv3d & \multicolumn{5}{|c|}{in\_ch: 5, out\_ch: 64, kernel: 3x3, stride: 1, pad: 1 } \\
\hline
\multirow{3}*{Time\_Embed} &  Linear & \multicolumn{5}{|c|}{in\_ch:64, out\_ch: 256 } \\
~ &  Activateion & \multicolumn{5}{|c|}{nn.SiLU} \\
~ &  Linear & \multicolumn{5}{|c|}{in\_ch:256, out\_ch: 256 } \\
\hline

\multirow{4}*{Downsample\_block\_1} &  ResnetBlock & \multicolumn{5}{|c|}{in\_ch:64, out\_ch: 64 } \\
~ &  ResnetBlock & \multicolumn{5}{|c|}{in\_ch:64, out\_ch: 64 } \\
~ &  Feature\_injetced\_from\_2D & \multicolumn{5}{|c|}{in\_ch:64, out\_ch: 64) } \\
~ &  Downsample(Conv3d) & \multicolumn{5}{|c|}{in\_ch:64, out\_ch: 64,kernel:3x3, stride:2) } \\
\hline
\multirow{3}*{Downsample\_block\_2} &  ResnetBlock & \multicolumn{5}{|c|}{in\_ch:64, out\_ch: 128 } \\
~ &  ResnetBlock & \multicolumn{5}{|c|}{in\_ch:128, out\_ch: 128 } \\
~ &  Feature\_injetced\_from\_2D & \multicolumn{5}{|c|}{in\_ch:128, out\_ch: 128) } \\
~ &  Downsample(Conv3d) & \multicolumn{5}{|c|}{in\_ch:128, out\_ch: 128,kernel:3x3, stride:2 } \\
\hline
\multirow{3}*{Downsample\_block\_3} &  ResnetBlock & \multicolumn{5}{|c|}{in\_ch:128, out\_ch: 192 } \\
~ &  ResnetBlock & \multicolumn{5}{|c|}{in\_ch:192, out\_ch: 192 } \\
~ &  Feature\_injetced\_from\_2D & \multicolumn{5}{|c|}{in\_ch:192, out\_ch: 192) } \\
~ &  Downsample(Conv3d) & \multicolumn{5}{|c|}{in\_ch:192, out\_ch: 192,kernel:3x3, stride:2 } \\
\hline
\multirow{2}*{Downsample\_block\_4} &  ResnetBlock & \multicolumn{5}{|c|}{in\_ch:192, out\_ch: 256 } \\
~ &  ResnetBlock & \multicolumn{5}{|c|}{in\_ch:256, out\_ch: 256 } \\
~ &  Feature\_injetced\_from\_2D & \multicolumn{5}{|c|}{in\_ch:256, out\_ch: 256) } \\
\hline
\multirow{3}*{Middle} &  ResnetBlock & \multicolumn{5}{|c|}{in\_ch:256, out\_ch: 256 } \\
~ &  ResnetBlock & \multicolumn{5}{|c|}{in\_ch:256, out\_ch: 256 } \\
\hline
\multirow{3}*{Upsample\_block\_1} &  ResnetBlock & \multicolumn{5}{|c|}{in\_ch:256, out\_ch: 256 } \\
~ &  ResnetBlock & \multicolumn{5}{|c|}{in\_ch:256, out\_ch: 256 } \\
~ &  Upsample & \multicolumn{5}{|c|}{Conv3d and F.interpolate} \\
\hline
\multirow{3}*{Upsample\_block\_2} &  ResnetBlock & \multicolumn{5}{|c|}{in\_ch:256, out\_ch: 192 } \\
~ &  ResnetBlock & \multicolumn{5}{|c|}{in\_ch:192, out\_ch: 192 } \\
~ &  Upsample & \multicolumn{5}{|c|}{Conv3d and F.interpolate} \\
\hline
\multirow{3}*{Upsample\_block\_3} &  ResnetBlock & \multicolumn{5}{|c|}{in\_ch:192, out\_ch: 128 } \\
~ &  ResnetBlock & \multicolumn{5}{|c|}{in\_ch:128, out\_ch: 128 } \\
~ &  Upsample & \multicolumn{5}{|c|}{Conv3d and F.interpolate} \\
\hline
\multirow{2}*{Upsample\_block\_4} &  ResnetBlock & \multicolumn{5}{|c|}{in\_ch:128, out\_ch: 64 } \\
~ &  ResnetBlock & \multicolumn{5}{|c|}{in\_ch:64, out\_ch: 64 } \\
\hline

\multirow{3}*{Out} &  Normalize & \multicolumn{5}{|c|}{64} \\

~ &  Activation & \multicolumn{5}{|c|}{nn.SiLU} \\

~ &  Conv3d & \multicolumn{5}{|c|}{in\_ch:64, out\_ch: 2, kernel: 3x3, stride: 1, pad: 1} \\
\hline

\end{tabular}
\label{tab:dif3D}
\end{table*}

\begin{table*}[!t]\centering
\caption{Architecture for the smaller variant of 3D diffusion model. Again, each ResnetBlock consists of 2 conv3D layers of the same channel and a skip connection. All ResnetBlocks are used with time embedding with an embedding layer, as well as a gradient checkpoint. We used a smaller number of channels for each layer and omitted the feature injection from 2D}
\begin{tabular}{|c|c|c|c|c|c|c|}
\hline
\multicolumn{2}{|c|}{layers} & \multicolumn{5}{|c|}{parameters} \\
\hline
input & Conv3d & \multicolumn{5}{|c|}{in\_ch: 5, out\_ch: 32, kernel: 3x3, stride: 1, pad: 1 } \\
\hline
\multirow{3}*{Time\_Embed} &  Linear & \multicolumn{5}{|c|}{in\_ch:32, out\_ch: 128 } \\
~ &  Activateion & \multicolumn{5}{|c|}{nn.SiLU} \\
~ &  Linear & \multicolumn{5}{|c|}{in\_ch:128, out\_ch: 128 } \\
\hline

\multirow{4}*{Downsample\_block\_1} &  ResnetBlock & \multicolumn{5}{|c|}{in\_ch:32, out\_ch: 32 } \\
~ &  ResnetBlock & \multicolumn{5}{|c|}{in\_ch:32, out\_ch: 32 } \\
~ &  Downsample(Conv3d) & \multicolumn{5}{|c|}{in\_ch:32, out\_ch: 32,kernel:3x3, stride:2) } \\
\hline
\multirow{3}*{Downsample\_block\_2} &  ResnetBlock & \multicolumn{5}{|c|}{in\_ch:32, out\_ch: 64 } \\
~ &  ResnetBlock & \multicolumn{5}{|c|}{in\_ch:64, out\_ch: 64 } \\
~ &  Downsample(Conv3d) & \multicolumn{5}{|c|}{in\_ch:64, out\_ch: 64,kernel:3x3, stride:2 } \\
\hline
\multirow{3}*{Downsample\_block\_3} &  ResnetBlock & \multicolumn{5}{|c|}{in\_ch:64, out\_ch: 64 } \\
~ &  ResnetBlock & \multicolumn{5}{|c|}{in\_ch:64, out\_ch: 64 } \\
~ &  Downsample(Conv3d) & \multicolumn{5}{|c|}{in\_ch:64, out\_ch: 64,kernel:3x3, stride:2 } \\
\hline
\multirow{2}*{Downsample\_block\_4} &  ResnetBlock & \multicolumn{5}{|c|}{in\_ch:63, out\_ch: 128 } \\
~ &  ResnetBlock & \multicolumn{5}{|c|}{in\_ch:128, out\_ch: 128 } \\
\hline
\multirow{3}*{Middle} &  ResnetBlock & \multicolumn{5}{|c|}{in\_ch:128, out\_ch: 128 } \\
~ &  ResnetBlock & \multicolumn{5}{|c|}{in\_ch:128, out\_ch: 128 } \\
\hline
\multirow{3}*{Upsample\_block\_1} &  ResnetBlock & \multicolumn{5}{|c|}{in\_ch:128, out\_ch: 128 } \\
~ &  ResnetBlock & \multicolumn{5}{|c|}{in\_ch:128, out\_ch: 128 } \\
~ &  Upsample & \multicolumn{5}{|c|}{Conv3d and F.interpolate} \\
\hline
\multirow{3}*{Upsample\_block\_2} &  ResnetBlock & \multicolumn{5}{|c|}{in\_ch:128, out\_ch: 64 } \\
~ &  ResnetBlock & \multicolumn{5}{|c|}{in\_ch:64, out\_ch: 64 } \\
~ &  Upsample & \multicolumn{5}{|c|}{Conv3d and F.interpolate} \\
\hline
\multirow{3}*{Upsample\_block\_3} &  ResnetBlock & \multicolumn{5}{|c|}{in\_ch:64, out\_ch: 64 } \\
~ &  ResnetBlock & \multicolumn{5}{|c|}{in\_ch:64, out\_ch: 64 } \\
~ &  Upsample & \multicolumn{5}{|c|}{Conv3d and F.interpolate} \\
\hline
\multirow{2}*{Upsample\_block\_4} &  ResnetBlock & \multicolumn{5}{|c|}{in\_ch:64, out\_ch: 32 } \\
~ &  ResnetBlock & \multicolumn{5}{|c|}{in\_ch:32, out\_ch: 32 } \\
\hline

\multirow{3}*{Out} &  Normalize & \multicolumn{5}{|c|}{64} \\

~ &  Activation & \multicolumn{5}{|c|}{nn.SiLU} \\

~ &  Conv3d & \multicolumn{5}{|c|}{in\_ch:32, out\_ch: 2, kernel: 3x3, stride: 1, pad: 1} \\
\hline

\end{tabular}
\label{tab:small}
\end{table*}

\end{document}